%% file: acl_latex.tex
\pdfoutput=1
\documentclass[11pt]{article}

\usepackage[preprint]{acl}

\usepackage{times}
\usepackage{latexsym}

\usepackage[T1]{fontenc}
\usepackage[utf8]{inputenc}

\usepackage{microtype}

\usepackage{inconsolata}

\usepackage{graphicx}

\usepackage[utf8]{inputenc} % allow utf-8 input
\usepackage[T1]{fontenc}    % use 8-bit T1 fonts
\usepackage{url}            % simple URL typesetting
\usepackage{booktabs}       % professional-quality tables
\usepackage{amsfonts}       % blackboard math symbols
\usepackage{nicefrac}       % compact symbols for 1/2, etc.
\usepackage{microtype}      % microtypography
\usepackage{xcolor}         % colors
\usepackage{multirow}
\usepackage{graphicx}

\newcommand{\fref}[1]{Figure~\ref{#1}}
\newcommand{\tref}[1]{Table~\ref{#1}}
\newcommand{\sref}[1]{\S\ref{#1}}

\usepackage{microtype}
\usepackage{graphicx}
\usepackage{amsmath}
\usepackage{bbm}
\usepackage{tcolorbox}

\usepackage{enumitem}
\usepackage{CJKutf8}

\usepackage[]{todonotes}

\title{On the Fluid Intelligence Evaluation for LLMs}
\title{Understanding LLMs' Fluid Intelligence Deficiency: \\ An Analysis of the ARC Task}

\newcommand{\authorsep}{\quad}

\author{
Junjie Wu$^1$\authorsep
Mo Yu$^2$\thanks{Co-corresponding authors.}\authorsep
Lemao Liu$^2$\authorsep
Dit-Yan Yeung$^1$\footnotemark[1]\authorsep
Jie Zhou$^2$\authorsep
\\
\textsuperscript{1}Hong Kong University of Science and Technology\\
\textsuperscript{2}WeChat AI, Tencent\\
\texttt{junjie.wu@connect.ust.hk} \quad \texttt{moyumyu@global.tencent.com} \\ \texttt{\{redmondliu, withtomzhou\}@tencent.com} \quad \texttt{dyyeung@ust.hk}
}

\begin{document}
\maketitle
\begin{abstract}
While LLMs have exhibited strong performance on various NLP tasks, it is noteworthy that most of these tasks rely on utilizing the vast amount of knowledge encoded in LLMs' parameters, rather than solving new problems without prior knowledge. In cognitive research, the latter ability is referred to as fluid intelligence, which is considered to be critical for assessing human intelligence. Recent research on fluid intelligence assessments has highlighted significant deficiencies in LLMs' abilities. In this paper, we analyze the challenges LLMs face in demonstrating fluid intelligence through controlled experiments, using the most representative ARC task as an example.
Our study revealed three major limitations in existing LLMs: limited ability for skill composition, unfamiliarity with abstract input formats, and the intrinsic deficiency of left-to-right decoding.
Our data and code can be found in \url{https://wujunjie1998.github.io/araoc-benchmark.github.io/}.

\end{abstract}

\newcommand{\lemao}[1]{\textcolor{red}{\textbf{#1 --Lemao}}}

\input{sections/introduction.tex}

\input{sections/results_on_arc.tex}

\input{sections/Investigate_on_ARAOC.tex}

\input{sections/factor.tex}

\input{sections/input.tex}

\input{sections/model.tex}

\input{sections/related_work.tex}

\section{Conclusion}
This paper presents an in-depth study of LLMs' fluid intelligence deficiencies using the ARC tasks, with a series of controlled experiments from multiple perspectives. %and uncovering several key findings. 
Through task decomposition, we introduce the atomic ARAOC benchmark, revealing that LLMs struggle with atomic operations despite their simplicity for humans. We further demonstrate that LLMs' task composition abilities are limited, as improvements on the decomposed ARAOC tasks via fine-tuning do not lead to better performance on ARC tasks. Additionally, our study shows that LLMs' difficulty in encoding abstract input formats is a major obstacle in addressing ARC tasks. Lastly, it shows an intrinsic limitation in the left-to-right paradigm of LLMs, which hinders their ability to achieve advanced fluid intelligence.

\section*{Limitations}
Due to the experiment budget, on all the ARC related experiments, we only evaluate LLMs on 100 tasks rather than the whole corpus following~\citet{wang2023hypothesis}, which may lead to potential bias in the evaluation results. Also, although most of the ARC tasks can be composed by the six atomic operations proposed by our work, there may still exist very few tasks that cannot be composed by our atomic operations, which may also introducing few bias to~\tref{tab:fine-tune arc performance}. We will try to provide more comprehensive results in future works once we get more experimental budgets, and propose more atomic operations that could be used to cover more ARC tasks.

%\section*{Ethical Considerations}
%Since this paper includes many responses generated by LLMs, it is possible that these LLM generated contents include toxic and harmful parts, requiring users to perform comprehensive data processing if they want to use our methods.

\section*{Acknowledgment}
This work has been made possible by a Research Impact Fund project (RIF R6003-21) and a General Research Fund project (GRF 16203224) funded by the Research Grants Council (RGC) of the Hong Kong Government.

\bibliography{custom}

\appendix
\clearpage

\input{sections/appendix.tex}

\end{document}

%% file: sections/introduction.tex
\begin{table*}[tb]
  \renewcommand\arraystretch{1.1}
  \centering
  \setlength{\tabcolsep}{1.5mm}
  \small
  \begin{tabular}{p{3cm}|p{3.5cm}|p{3.5cm}|p{4.3cm}}
    \toprule[1pt]
    \multicolumn{1}{c}{\textbf{II}~\cite{honovich2023instruction}} & \multicolumn{1}{c}{\textbf{Deer}~\cite{yang2024language}} & \multicolumn{1}{c}{\textbf{Mini Scan}~\cite{qiuphenomenal}} & \multicolumn{1}{c}{\textbf{ARC}~\cite{chollet2019measure}} \\
    \midrule[0.5pt]

    \begin{minipage}{3cm}
    \vspace{-4mm}
      \footnotesize
      \textbf{Input}: turn \newline
      \textbf{Output}: play \newline
      \newline
      \textbf{Input}: floor \newline
      \textbf{Output}: level \newline
      \newline
      \textbf{Input}: embrace \newline
      \textbf{Output}: cover \newline
      ......
      %%Output: communicate \newline
      %%Input: wash \newline
      %Output: washing
      %\newline
      \newline
      \textbf{Instruction: ?}
    \end{minipage}
    &
    \begin{minipage}{3.5cm}
      \footnotesize
      \textbf{Rule Type}: \newline
      There exists \_ , which \_. \newline
      \newline
      \textbf{Fact 1}: \newline
      Crabs are generally covered with a thick exoskeleton. \newline
      \newline
      \textbf{Fact 2}: \newline
      Lobsters are invertebrates with a hard protective exoskeleton. \newline
      %\newline
      %Fact3: The mollusc shell is typically a calcareous exoskeleton which encloses, supports, and protects the soft parts of an animal. \newline
        %\textbf{Fact 3}: \newline 
        ......
      \newline
      \textbf{Rule: ?}
    \end{minipage}
    &\begin{minipage}{3cm}
      \vspace{-3mm} % This reduces the space between the top of the page and the image
      \centering
      \includegraphics[width=3cm]{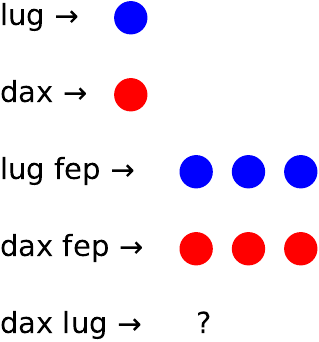}
    \end{minipage}
    & 
    \begin{minipage}{4.3cm}
      \vspace{-3mm} % This reduces the space between the top of the page and the image
      \centering
      \includegraphics[width=4.3cm]{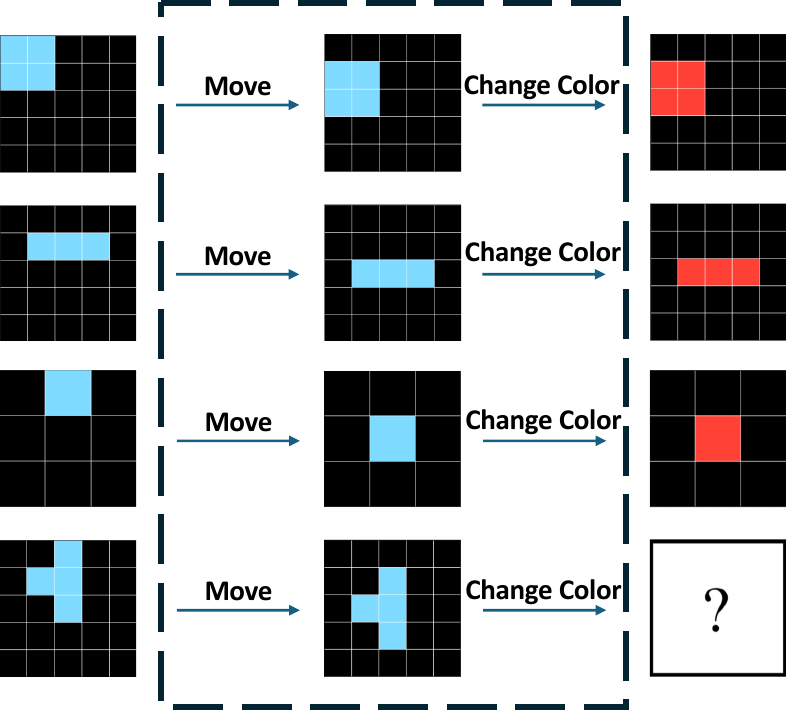}
    \end{minipage}
    \\
    \midrule[0.5pt]
    \begin{minipage}{3cm}
      \centering
      \includegraphics[width=2.5cm]{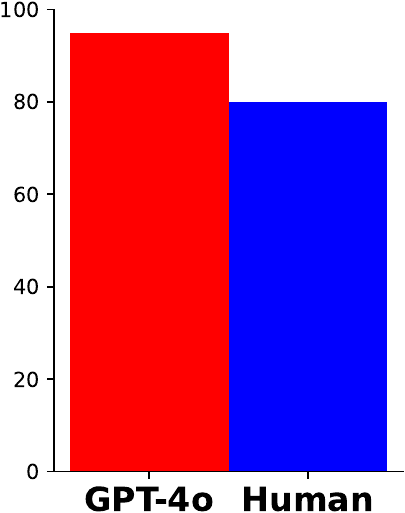}
    \end{minipage} &
    \begin{minipage}{3.5cm}
      \centering
      \includegraphics[width=2.5cm]{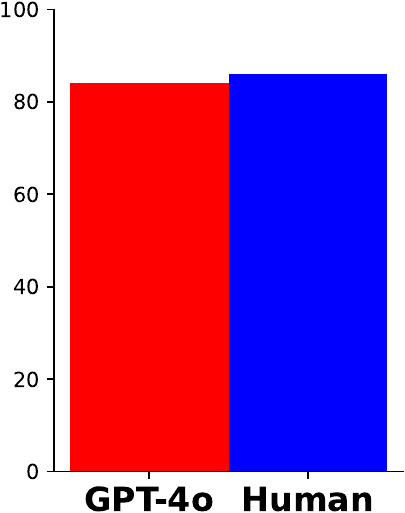}
    \end{minipage} &
    \begin{minipage}{3cm}
      \centering
      \includegraphics[width=2.5cm]{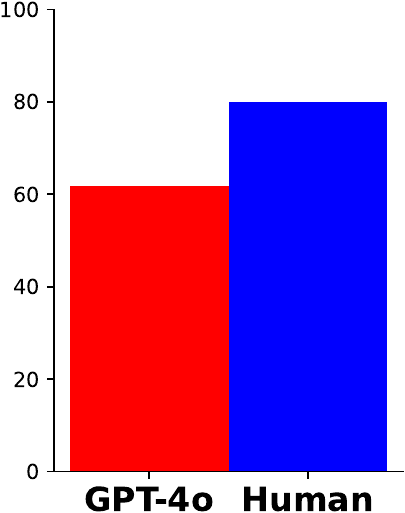}
    \end{minipage} &
    \begin{minipage}{5cm}
      \centering
      \includegraphics[width=2.5cm]{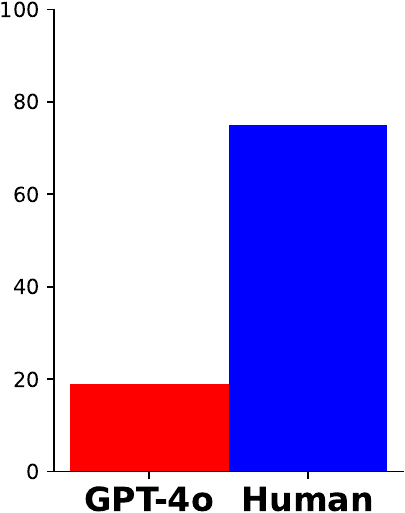}
    \end{minipage} \\
\bottomrule[1pt]
  \end{tabular}
  \caption{Examples of different inductive reasoning tasks with GPT-4o and human's scores. Check Appendix~\ref{appendix:inductive examples} for details of score calculation on the first three tasks. The transformation rule of the ARC example can be decomposed into two atomic operations in the dotted box %\lemao{Add more information about task composition.}
    %Two decomposed atomic operations 
    %described in~\tref{tab:atom operations}
    :1) move down the blue subgrid for one step; 2) change its color to red.}
  \vspace{-0.2in}
  \label{tab:inductive reasoning examples}
\end{table*}

\section{Introduction}
\label{intro}
Large language models (LLMs) have demonstrated impressive performance on a range of challenging NLP tasks~\cite{davis2023benchmarks, zhao2024large, wang2023hypothesis, yang2024language, frieder2024mathematical}, %These successes 
which naturally leads to the question: \emph{how close are LLMs to achieving human-level intelligence?}

To explore this question, it is useful to draw on established research in human intelligence~\cite{cattell1963theory, cattell1971abilities}, which categorizes intelligence into two major types: \textbf{crystallized intelligence}, the ability to apply prior knowledge to solve problems, and \textbf{fluid intelligence}, the ability to tackle new problems without relying on pre-existing knowledge. Of the two, fluid intelligence is often viewed as more indicative of general cognitive ability~\cite{jaeggi2008improving, chollet2019measure, barak2024investigating}, as it captures the capability to solve novel problems. %As a result, most widely used intelligence assessments focus on tasks involving fluid intelligence. 
Moreover, evaluating fluid intelligence is essential for assessing the reasoning abilities of LLMs since LLMs have been exposed to and memorized vast amounts of knowledge during pre-training~\cite{kaplan2020scaling, biderman2024emergent}, potentially blurring the line between their reasoning ability and memorization capacity.

Drawing from both cognitive~\cite{jensen1998factor} and AI research~\cite{chollet2019measure, barak2024investigating}, an ideal approach to evaluate fluid intelligence in LLMs involves evaluating their ability to perform abstract inductive reasoning, i.e., induct a general pattern solely from given input-output examples and apply this pattern to deduce the correct outputs for new inputs. %By incorporating patterns that are either counterfactual or highly infrequent, this evaluation could ensure that LLMs must rely on their reasoning capabilities rather than pre-existing knowledge.
The Abstraction and Reasoning Challenge (ARC)~\cite{chollet2019measure}, which requires models to induct transformation rules from input-output grid pairs (as shown in \tref{tab:inductive reasoning examples}), is a benchmark well-suited for this purpose. Due to the abstract nature of ARC tasks, LLMs cannot rely on memorization or external knowledge to solve them. In contrast, many existing inductive reasoning tasks~\cite{honovich2023instruction, yang2024language, qiuphenomenal} fail to prevent the use of memorization shortcuts, making those tasks easier for LLMs to solve, as shown in~\tref{tab:inductive reasoning examples}.

Therefore, the ARC task has become the de facto standard for measuring machine fluid intelligence, sparking a wave of recent studies aimed at improving LLM performance on it~\cite{acquaviva2022communicating, xullms, wang2023hypothesis, wang2024speak}. Despite these efforts, LLMs continue to struggle with the ARC task. For example, even the state-of-the-art GPT-4o with careful prompting can only correctly solve 19\% of tasks (see~\tref{tab:arc performance}), which falls far short of the average human performance of $\sim$75\%~\cite{legris2024h}. These observations lead us to explore a fundamental question: {\em why is the ARC task so challenging for LLMs?}

\begin{table}[tb]
\small
\centering
\setlength{\tabcolsep}{2mm}
\begin{tabular}{lcc}
\toprule
\textbf{Input Format} & \textbf{Output Format} 
%& \textbf{$\text{P-Acc}_{\text{A}}$}$\uparrow$ & \textbf{$\text{P-Acc}_{\text{M}}$}$\uparrow$ 
& \textbf{Correct Num}$\uparrow$ \\
\midrule
Visual & Visual & 0 \\
Visual & Textual & 1\\
Visual + Textual & Visual & 0\\
Visual + Textual & Textual & 16 \\
Textual & Textual & 19 \\

\bottomrule
\end{tabular}
\caption{Evaluation results of GPT-4o on the 100 ARC tasks using different input/output formats.}

\vspace{-0.1in}
\label{tab:different format}
\end{table}

To this end, this paper proposes to investigate the answers to the above question from multiple perspectives.
Our first intuition is inspired by an observation that the transformation rule of each ARC task can be regarded as a composition of atomic operations (e.g., the transformation rule in~\tref{tab:inductive reasoning examples} can be split into two atomic operations).
This motivates us to study the ARC tasks through task decomposition. Hence, we first decompose ARC tasks into atomic operations as transformation rules to construct a benchmark \textbf{A}bstraction and \textbf{R}easoning on \textbf{A}tom \textbf{O}peration  \textbf{C}orpus (\textbf{ARAOC}). However, LLMs perform poorly on some atomic operations on ARAOC, while this task is trivial to humans (\sref{sec: atom operation}), which indicates that their fluid intelligences are limited. Then, from a perspective of task decomposition, we %conduct experiments to 
evaluate the composition ability for LLMs on both ARAOC and ARC benchmarks. Our finding reveals that the limited composition ability for LLMs also contributes to their failures on fluid intelligence evaluation tasks (\sref{sec:factor}).

Next, we explore the challenge from the abstract representation format perspective. Our another intuition is that LLMs may lack the ability to understand two-dimensional NumPy arrays (matrices) that are commonly used to represent the 2D pixel grid inputs in ARC and ARAOC tasks~\cite{xullms,wang2023hypothesis}, which hinders their performances. We thereby design experiments to investigate whether LLMs understand such matrices-form inputs, and also convert matrices into natural language to see whether it enhances LLM's performances (\sref{sec:matrix}). 
Finally, we investigate the challenge from the modeling perspective. We conduct experiments to analyze the effect of left-to-right autoregressive decoding on model performances and analyze whether LLMs could correctly utilize important information on ARAOC tasks (\sref{sec:model}). Extensive experimental results in~\sref{sec:matrix} and~\sref{sec:model} give us several hints on why LLMs cannot perform ARC and ARAOC tasks well, which further motivating us to design strategies to enhance their corresponding capabilities. 
Overall, the contributions of this paper are summarized as follows:
\begin{enumerate}[noitemsep,nolistsep,leftmargin=*]
    \item This paper makes an initial attempt to study fluid intelligence of LLMs using the ARC tasks and conduct an in-depth study from multiple perspectives.
    \item We propose the ARAOC benchmark that assesses the fluid intelligence over atomic operations from ARC, which is extremely simple to humans yet surprisingly challenging for LLMs.
    % as a decomposed version of ARC, which is extremely simple to humans yet surprisingly challenging for LLMs as a good testbed for inductive reasoning.
    \item We obtain several valuable findings through controllable experiments on ARAOC and ARC, and reveal the challenges of LLMs on internal factors, task composition, input format as well as modeling with left-to-right Transformer. 
\end{enumerate}

%% file: sections/results_on_arc.tex
\section{Evaluating Fluid Intelligence on ARC}
\label{evaluate llm on arc}

\subsection{ARC Benchmark}
\label{sec:arc setting}
%\paragraph{ARC Benchmark.}
We start by evaluating the fluid intelligences of existing LLMs using the ARC benchmark, which comprises 400 training and 400 evaluation tasks. As shown in~\tref{tab:inductive reasoning examples}, each ARC task includes several 2D input-output grid pairs that define a unique transformation rule, with each grid ranging from $1 \times 1$ to $30 \times 30$ pixels, and each pixel being one of ten colors (see~\fref{fig:original prompt} for the names of the ten colors). An LLM must induct the transformation rule from the given input-output grid pairs and use it to predict the output grid for a testing input grid. Due to the high cost of closed-source LLMs, we follow~\citet{wang2023hypothesis} and use a subset of 100 training tasks in ARC for evaluation~\footnote{\scriptsize{Additionally, we evaluated GPT-4o on all 400 training tasks, where it achieved an Acc score of 18.50. This result aligns with the score reported in~\tref{tab:arc performance}, further supporting the representativeness of the subset.
}}.

\begin{table}[tb]
\small
\centering
\setlength{\tabcolsep}{4mm}
\begin{tabular}{lcc}
\toprule
\textbf{LLM} & \textbf{Acc}$\uparrow$ 
%& \textbf{$\text{P-Acc}_{\text{A}}$}$\uparrow$ & \textbf{$\text{P-Acc}_{\text{M}}$}$\uparrow$ 
& \textbf{Not M\%}$\downarrow$ \\
\midrule
Mistral & 2.00 
%& 32.59 & 62.67 
& 48.00 \\
Llama-3 & 5.00 
%& 49.56 & 73.98 
& 33.00 \\
\midrule
$\text{Mistral-FT}_{\text{ARC}}$ & 3.00 
%&44.74 &67.79 
&34.00 \\
$\text{Llama-3-FT}_{\text{ARC}}$ & 9.00
%& 54.20& 76.34
& 29.00\\
\midrule
$\text{Mistral-8*7B}$ &3.00
%&44.74 &67.79 
& 27.00\\
$\text{Llama-3-70B}$ &9.00
%& 54.20& 76.34
& 24.00\\
\midrule
GPT-3.5 & 6.00 
%& 46.38 & 71.35
& 35.00 \\
%GPT-4 & \textbf{17.00} 
%& \textbf{69.52} & \textbf{82.76} 
%& \textbf{16.00} \\
GPT-4o & \textbf{19.00} &\textbf{11.00} \\
\midrule
GPT-o1* & 18.00 & 10.00 \\
\bottomrule
\end{tabular}
\caption{Evaluation results on the 100 ARC tasks, where Acc %$\text{P-Acc}_{\text{A}}$, and $\text{P-Acc}_{\text{M}}$ 
is represented as percentages. $\text{FT}_{\text{ARC}}$ denotes fine-tuning on ARC tasks. The best results in each column are \textbf{boldfaced}. *GPT-o1 is evaluated on a partial subset, where GPT-4o obtains \emph{16.00} and \emph{10.00} for both scores.
}
\vspace{-0.2in}
\label{tab:arc performance}
\end{table}

\begin{table*}[tb]
  \renewcommand\arraystretch{1.1}
  \centering
  \setlength{\tabcolsep}{2mm}
  \small
  \begin{tabular}{p{2cm}p{7cm}p{5cm}}
    \toprule[1pt]
   
     \textbf{Name} & \textbf{Description/Transformation Rule} & \textbf{Example} \\
     \midrule[0.5pt]
    \textbf{Move} & Move a subgrid in the input grid for several steps towards a single direction in one of \{Up, Down, Left, Right, Up-left, Up-right, Down-left, Down-right\} to form the output grid. Note that the moved subgrid could not surpass the boundary of the input grid. & \begin{center}\vspace{-1mm}\includegraphics[width=5cm]{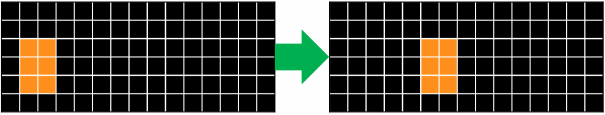}\vspace{-3mm}\end{center} \\
    \midrule[0.5pt]
   \textbf{Change Color} & Change the color of a subgrid in the input grid to another color other than black to form the output grid. & \vspace{-3.5mm} \begin{center}\includegraphics[width=5cm]{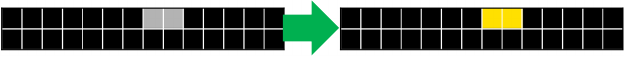}\end{center}\vspace{-5mm} \\
   \midrule[0.5pt]
   \textbf{Copy} & Copy a subgrid in the input grid and move it with Move to form the output grid, while making sure that the copied subgrid could neither surpass the boundary of the input grid, nor overlap with the original subgrid. & \vspace{-3.5mm} \begin{center}\includegraphics[width=5cm]{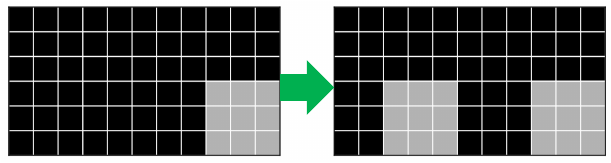}\end{center} \vspace{-5.5mm} \\
   \midrule[0.5pt]
   \textbf{Mirror} & Mirror the input grid towards a single direction in one of \{Up, Down, Left, Right\} to form the output grid. & \vspace{-4mm} \begin{center}\includegraphics[width=5cm]{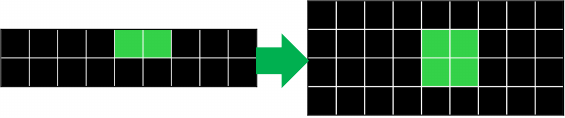}\end{center} \vspace{-5.5mm}\\
   \midrule[0.5pt]
   \textbf{Fill Internal} & The input grid has a closed subgrid whose internal is black. Fill the internal black part of this subgrid with another color to form the output grid. &\vspace{-3mm} \begin{center}\includegraphics[width=5cm]{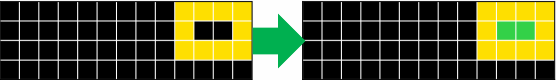}\end{center} \vspace{-6.5mm} \\
   \midrule[0.5pt]
   \textbf{Scale} & Some pixels in the input grid are colored with a specific color. Let the number of rows and columns of the input grid be \(a\) and \(b\), respectively. First, the input grid will be copied \(a \times b\) times. These copies will then be arranged in an output grid with \(a \times a\) rows and \(b \times b\) columns, placed from top to bottom and left to right. Finally, if the position \((i, j)\) in the input grid is black, the \(i \times j\)-th copy in the output grid will be converted to black. & \vspace{-4.5mm} \begin{center}\includegraphics[width=5cm]{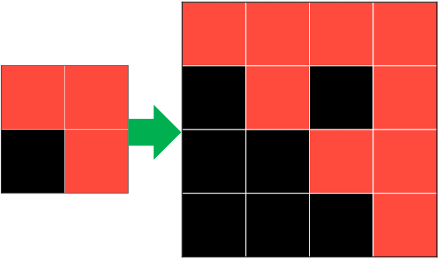}\end{center} \vspace{-6mm} \\
    \bottomrule[1pt]
  \end{tabular}
  \caption{Descriptions and examples of the six atomic operations we use in this paper.}
  \vspace{-0.1in}
  \label{tab:atom operations}
\end{table*}

\begin{table*}[tb]
\renewcommand\arraystretch{1.1}
\centering
\setlength{\tabcolsep}{0.8mm}
\small
\begin{tabular}{lcc|cc|cc|cc|cc|cc}
\toprule[1pt]
\multirow{2}*{LLM} & \multicolumn{2}{c}{\textbf{Move}} & \multicolumn{2}{c}{\textbf{Change Color}} & \multicolumn{2}{c}{\textbf{Copy}} & \multicolumn{2}{c}{\textbf{Mirror}} & \multicolumn{2}{c}{\textbf{Fill Internal}} & \multicolumn{2}{c}{\textbf{Scale}} \\
 & Acc$\uparrow$ & Not M\%$\downarrow$ & Acc$\uparrow$ & Not M\%$\downarrow$ & Acc$\uparrow$ & Not M\%$\downarrow$ & Acc$\uparrow$ & Not M\%$\downarrow$ & Acc$\uparrow$ & Not M\%$\downarrow$ & Acc$\uparrow$ & Not M\%$\downarrow$ \\
\midrule[0.5pt]
Mistral & 2.00 & 36.00 & 15.00 & 30.00 & 2.00 & 43.00 & 1.00 & 97.00 & 9.00 & 31.00 & 0.00 & 98.00 \\
Llama-3 & 1.00 & 19.00 & 39.00 & 17.00 & 4.00 & 13.00 & 2.00 & 96.00 & 63.00 & 6.00 & 1.00 & 89.00 \\
\midrule
Mistral-8*7B &2.00&10.00&57.00&5.00&2.00&7.00&5.00&95.00&50.00&3.00&\textbf{3.00}&81.00 \\
Llama-3-70B &8.00&15.00&92.00&1.00&4.00&11.00&7.00&75.00&64.00&3.00&\textbf{3.00}&80.00 \\
\midrule
GPT-3.5 & 4.00 & 27.00 & 48.00 & 13.00 & 4.00 & 29.00 & 6.00 & 89.00 & 58.00 & 12.00 & 1.00 & 80.00 \\
%GPT-4 & \textbf{14.00} & \textbf{3.00} & \textbf{97.00} & \textbf{0.00} & \textbf{13.00} & \textbf{6.00} & \textbf{14.00} & \textbf{52.00} & \textbf{100.00} & \textbf{0.00} & \textbf{3.00} & \textbf{70.00} \\
GPT-4o &\textbf{13.00}&\textbf{0.00}&\textbf{98.00}&\textbf{0.00}&\textbf{15.00}&\textbf{0.00}&\textbf{12.00}&\textbf{48.00}&\textbf{96.00}&\textbf{0.00}&2.00&\textbf{72.00} \\
%\midrule
%$\text{Mistral-FT}_{\text{Atom}}$ & 12.00 & 11.00 & 100.00 & 0.00 & 20.00 & 6.00 & 26.00 & 52.00 & 97.00 & 2.00 & 89.00 & 0.00 \\
%$\text{Llama-3-FT}_{\text{Atom}}$ & 13.00 & 9.00 & 98.00 & 1.00 & 14.00 & 8.00 & 27.00 & 54.00 & 97.00 & 1.00 & 78.00 & 99.08 \\
\bottomrule[1pt]
\end{tabular}
\caption{Results on ARAOC. %$\text{FT}_{\text{Atom}}$ denotes fine-tuning on atomic operation data. 
Acc is shown in percentage. The best results under each column are \textbf{boldfaced}.}
\vspace{-0.2in}
\label{tab:araoc results}
\end{table*}

\subsection{Comparing Text- and Visual-Based LLMs}
Since ARC tasks are presented in a 2D visual grid format, we can employ both visual-based LLMs (\textbf{Visual}) and text-based LLMs through
%Since LLMs cannot process visual inputs directly, 
converting input-output grids into matrices represented by NumPy arrays following existing works~\cite{xullms,wang2023hypothesis} (\textbf{Textual}). Therefore, we first investigate the performances of these two types of LLMs on ARC by prompting GPT-4o with 5 different input-output formats (check Appendix~\ref{appendix:prompts} for the prompts). As shown in~\tref{tab:different format}, prompting GPT-4o solely with textual input-output format yielding the best performance on the 100 ARC tasks. On the other hand, it seems extremely challenging for visual-based LLMs to finish ARC tasks, where we provide detailed analysis in Appendix~\ref{appendix:visual analysis}. \textbf{Based on the results, we apply the textual only input-output format and refer \textit{``LLMs''} to text-based LLMs in the rest of the paper.}
%, as illustrated in~\fref{fig:example matrix input}.

\subsection{Comparing Different LLMs on ARC}
\label{sec:evaluate on original arc}
\paragraph{Evaluated LLMs.}
\label{evaluated llms}
We evaluate both closed-source and open-source LLMs. For closed-source models, we use GPT-4o and GPT-3.5. For open-source LLMs, we select Mistral (\texttt{Mistral-7B-Instruct-v0.2})~\cite{jiang2023mistral} and Llama-3 (\texttt{Llama-3-8B-Instruct})~\cite{llama3}. Additionally, we include the recently released GPT-o1 (\texttt{o1-preview}) model, known for its strong reasoning abilities, for comparison. Due to the slow inference speed and limited quota of GPT-o1, we evaluate it on a subset of 50 tasks and report the performance of both GPT-4o and GPT-o1 on this subset. Check Appendix~\ref{appendix:inference config} for details on the inference configurations.

%For all the models, we maintain their official prompt templates unchanged and the inference configurations are listed in Appendix~\ref{appendix:inference config}.

\paragraph{Evaluation Metrics.}
The primary metric we use to evaluate the performance of LLMs is the accuracy of their predictions (Acc). Additionally, since we observe that the shape of the LLMs' predicted output grids does not always align with the ground truth, we report the percentage of mismatched predictions for each LLM (Not M\%), where lower scores indicate better performance.

\paragraph{Results.}
The evaluation results are presented in~\tref{tab:arc performance}. We observe that, although GPT-4o significantly outperforms other LLMs, its performance remains far from ideal. Moreover, GPT-o1 shows almost no improvement over GPT-4o on the evaluated subset. Hence, due to its low speed and limited quota, we do not include GPT-o1 in the following experiments. 

For the other LLMs, handling ARC tasks seems extremely challenging, with more than one-third of their predictions failing to match the shape of the corresponding ground truth. To examine the impact of model size on ARC performance, we further experiment with Mistral-8*7B (\texttt{Mixtral-8x7B-Instruct-v0.1}) and Llama-3-70B (\texttt{Llama-3-70B-Instruct}). As shown in~\tref{tab:arc performance}, larger LLMs consistently outperform smaller ones across all tasks, indicating that models with more parameters exhibit stronger fluid intelligence on ARC tasks. However, their overall performance remains poor. We hypothesize that this poor performance is due to the LLMs' unfamiliarity with the style of these tasks. Consequently, we further fine-tuned Mistral and Llama-3 on a separate ARC evaluation set that do not overlap with the 100 ARC tasks used in~\tref{tab:arc performance} using LoRA~\cite{hu2021lora}, and evaluated them on the 100 ARC tasks (check fine-tuning details in Appendix~\ref{appendix:lora}). However, as shown in~\tref{tab:arc performance}, even though fine-tuning on ARC tasks improves the LLMs' performance, the results remain suboptimal, with Acc scores below 10\%.

In summary, these experiments demonstrate the significant challenge LLMs face in successfully completing ARC tasks, motivating us to further investigate the underlying reasons for this difficulty.

%\lemao{Please insert finetuning experiments into this subsection to further highlight the challenge of ARC tasks for LLMs.}

%% file: sections/Investigate_on_ARAOC.tex
\section{Breaking ARC into Atomic Operations}
\label{sec: atom operation}
%The poor performance of LLMs on ARC tasks motivates us to explore the reasons behind this issue. 
As mentioned in~\sref{intro}, the transformation rule of each ARC task can be decomposed into several atomic operations (e.g., the rule in~\tref{tab:inductive reasoning examples} can be broken into moving the subgrid and changing its color), which motivates us to analyze the challenges of LLMs from a task decomposition perspective. To this end, we first decompose the ARC tasks into simplified tasks and form the ARAOC benchmark that consists of various atomic operations, then use ARAOC to evaluate the fluid intelligence of LLMs.

\subsection{ARAOC Benchmark}

%\lemao{Remove the finetuning results from Table 3 and combine them into Table 6 to show performance gap on ARAOC and ARC, which demonstrates the composition challenge in sec 4.}

\label{sec:araoc benchmark}
To evaluate LLMs' fluid intelligence with atomic operations, we first manually go through all the tasks in ARC's training and evaluation sets, then conclude six atomic operations that can compose the transformation rules for most of the ARC tasks. Check~\tref{tab:atom operations} for atomic operations' descriptions. 

For each atomic operation, we use it as the transformation rule to build 100 tasks with 3 input-output training pairs and 1 testing pair, which follows the standard ARC setting (check Appendix~\ref{appendix:araoc} for the crafting details). This finally leads to a benchmark named \textbf{A}bstraction and \textbf{R}easoning on \textbf{A}tom \textbf{O}peration  \textbf{C}orpus (\textbf{ARAOC}) with 600 distinct tasks. %Specifically, for each task in ARAOC, we have three input-output grid pairs as the few-shot examples, and a single input grid that needs LLMs to infer its corresponding output grid. 
We evaluate all LLMs in~\sref{evaluated llms} on ARAOC and additionally include Mistral-8*7B and Llama-3-70B to study the impact of model size.

%list the results in~\tref{tab:araoc results}.

\iffalse
\begin{table}[tb]
\small
\centering
\setlength{\tabcolsep}{1mm}
\begin{tabular}{llcccc}
\toprule
& \textbf{COMB}& \textbf{Mistral}& \textbf{Llama-3} & \textbf{GPT-3.5}& \textbf{GPT-4o} \\
\midrule[0.5pt]
\multirow{6}{*}{\textbf{Move}}&Up 1 &0.00&12.00&0.00&26.00\\
&Up 2 &2.00&6.00&2.00&12.00\\
&Up 3 &4.00&4.00&0.00&8.00\\
\cmidrule{2-6}
&Up-right 1 &0.00&2.00&0.00&10.00\\
&Up-right 2 &0.00&0.00&0.00&0.00\\
&Up-right 3 &2.00&2.00&0.00&4.00\\
\midrule[0.5pt]
\multirow{6}{*}{\textbf{Copy}}&Up 1 &4.00&16.00&6.00&40.00\\
&Up 2 &8.00&10.00&6.00&26.00\\
&Up 3 &10.00&12.00&8.00&16.00\\
\cmidrule{2-6}
&Up-right 1 &2.00&8.00&0.00&16.00\\
&Up-right 2 &4.00&4.00&4.00&4.00\\
&Up-right 3 &2.00&4.00&0.00&2.00\\
\bottomrule
\end{tabular}
\caption{Further analysis results regarding Move and Copy. \textbf{COMB} is the abbreviation of combination. We only list Acc scores (in percentage) here for simplicity, and other metric scores are listed in Table Y.}
\label{tab:controllable}
\end{table}
\fi

\paragraph{Results.}
\label{sec:araoc results}
As shown in~\tref{tab:araoc results}, GPT-4o largely outperforms other LLMs across almost all tasks in the ARAOC benchmark, achieving nearly 100\% Acc scores on the Change Color and Fill Internal tasks, demonstrating its high fluid intelligence. Additionally, %GPT-3.5 and Llama-3 produce comparable results, 
Llama-3/Llama-3-70B outperforms Mistral/Mistral-8*7B, suggesting that pre-training %on higher-quality data 
with a greater number of parameters can enhance the fluid intelligence of LLMs. Also, similar to~\tref{tab:arc performance}, larger LLMs continue to outperform smaller ones across tasks, further illustrating the above point. However, all LLMs still encounter substantial difficulties with tasks related to Move, Copy, Mirror, and Scale, failing to predict the correct shapes of output grids for the latter two atomic operations on more than \textasciitilde50 tasks.

\subsection{Further Analysis}
%Moreover, the results in~\tref{tab:araoc results} %and~\tref{tab:composition} 
%show that LLMs' Acc scores on Move and Copy tasks in ARAOC are still less than 20\%, even they have been trained on similar data. 
\paragraph{Analysis I: Internal Factors.}
As concluded from~\sref{sec:araoc results}, all the LLMs exhibit poor performances on Move and Copy tasks in ARAOC. To analyze whether this is caused by the internal complexity of Move and Copy, we investigate factors that may affect the complexity of Move and Copy, and their influences on LLMs' performances. Given that Copy can actually be viewed as first copying the original subgrid, then moving the copied subgrid several steps in specific directions, we intuitively consider two factors in this study: 1) the number of steps the subgrid/copied subgrid moves; 2) the direction in which the subgrid/copied subgrid moves.

\paragraph{Setup.}
Specifically, we choose {Up, Up-right} and {1 step, 2 steps, 3 steps} as our candidate moving directions and steps, respectively. We then generate 50 input grids for each atomic operation, ensuring that these grids can be transformed into valid output grids based on any combination of the two candidate sets (e.g., Up for 1 step). For each input grid, we create 6 tasks corresponding to all 6 combinations of the candidate sets, and evaluate the closed-source (GPT-4o) and open-source (Llama-3) LLMs, which performed better in~\tref{tab:araoc results}, as representatives on these tasks.

\paragraph{Results.}
Results are shown in~\tref{tab:controllable}. We observe that 
%Mistral and GPT-3.5 can hardly finish the given tasks, and even scoring 0 under ``Up'', ``Up-right 1'' and ``Up-right 2'', rendering their results not indicative. 
%For GPT-4 and Llama-3, we notice that 
for both Move and Copy, a larger number of steps would lead to lower Acc scores. This could be because as the number of steps increases, LLMs need to focus on a longer context to induce the atomic operation, which leads to more challenges. Additionally, LLMs appear to be more adept with subgrids that move in a straight direction, as their performance on "Up"-related tasks is significantly higher than on "Up-right"-related tasks. Even when considering "Up-right 1" as a two-step move (one step "Up" and one step "Right"), LLMs' Acc scores on "Up-right 1" are still lower than those on "Up 2", further supporting our previous assertion. %Overall, we conclude that the inductive reasoning ability of LLMs can be influenced by various intrinsic factors related to different operations, which control the complexity of such atomic operations. %Future work should pay more attention to these factors when evaluating LLMs' inductive reasoning capabilities with atomic operations.

\begin{table}[tb]
\small
\centering
\setlength{\tabcolsep}{3mm}
\begin{tabular}{llcc}
\toprule
& \textbf{COMB}& \textbf{Llama-3} & \textbf{GPT-4o} \\
\midrule[0.5pt]
\multirow{6}{*}{\textbf{Move}}&Up 1 &12.00&24.00\\
&Up 2 &6.00&26.00\\
&Up 3 &4.00&17.00\\
\cmidrule{2-4}
&Up-right 1 &2.00&9.00\\
&Up-right 2 &0.00&2.00\\
&Up-right 3 &2.00&1.00\\
\midrule[0.5pt]
\multirow{6}{*}{\textbf{Copy}}&Up 1 &16.00&46.00\\
&Up 2 &10.00&38.00\\
&Up 3 &12.00&27.00\\
\cmidrule{2-4}
&Up-right 1 &8.00&11.00\\
&Up-right 2 &4.00&6.00\\
&Up-right 3 &4.00&10.00\\
\bottomrule
\end{tabular}
\caption{Analysis I's Acc scores. \textbf{COMB} refers to combination. See~\tref{tab:controllable_plus} for the Not M\% scores.}
\vspace{-0.2in}
\label{tab:controllable}
\end{table}

\paragraph{Analysis II: Effect on Input Size.}
%Another reason LLMs struggle with ARAOC may relate to the size of input grids, where larger input grids should bring more difficult tasks and vice versa. To test this hypothesis, 
We %evaluate LLMs on Move and Copy tasks for the effect on input size via 
evaluate LLMs on 100 Move and Copy tasks with smaller sizes (crafting details are included in Appendix~\ref{appendix:small size}). 
%and we randomly initialized the 100 tasks for each atomic operation from a range that is half of the original range listed in Appendix~\ref{appendix:araoc}. This results in 100 new tasks for both atomic operations, with an average size of $4.96 \times 4.89$ and $4.81 \times 4.80$, respectively. For comparison, the original average sizes are $10.07 \times 10.16$ and $9.72 \times 9.62$, respectively. 
The evaluation results on these tasks are listed in Table \ref{tab:large matrix}, where LLMs perform significantly better on Move and Copy tasks with smaller input sizes. This indicates that the size of matrix-format input does affect LLMs' understanding of ARAOC tasks and thus influences their performance on ARAOC. %See Appendix X for prompts used in this section.
%\mo{Conclusion: ARC is a natural long sequence understanding task (Combined with Conclusion 2 and 4.3)}

\textbf{Overall, this section shows that the performances of LLMs is largely affected by the superficial properties of the input grids, 
and LLMs fail to grasp the underlying concept of the operations. This result further suggests that LLMs rely more on pattern recognition and memorization, akin to crystallized intelligence, rather than reasoning through abstract, novel relationships (fluid intelligence).}
In the following, we provide further insights into LLMs' deficiencies through the lens of three challenges on ARC and ARAOC: task composition (\sref{sec:factor}), LLMs' understanding of task inputs (\sref{sec:matrix}), and LLMs' modeling strategies (\sref{sec:model}).

%% file: sections/factor.tex
\iffalse
\begin{table}[tb]
\small
\centering
\setlength{\tabcolsep}{0.7mm}
\begin{tabular}{lcccc}
\toprule
\textbf{LLM} & \textbf{Acc}$\uparrow$ & \textbf{$\text{P-Acc}_{\text{A}}$}$\uparrow$ & \textbf{$\text{P-Acc}_{\text{M}}$}$\uparrow$ & \textbf{Not M\%}$\downarrow$ \\
\midrule
%Mistral & 2.00 & 32.59 & 62.67 & 48.00 \\
%Llama-3 & 5.00 & 49.56 & 73.98 & 33.00 \\
%\midrule
$\text{Mistral-FT}_{\text{Atom}}$ &1.00 &42.46 &68.49 &38.00 \\
%$\text{Mistral-FT}_{\text{ARC+Atom}}$ & 6.00&51.31 &71.27 &28.00 \\
%\midrule
$\text{Llama-3-FT}_{\text{Atom}}$ & 2.00&42.77 &71.28 &40.00 \\
%$\text{Llama-3-FT}_{\text{ARC+Atom}}$ &6.00 &51.81 &74.02 &30.00 \\
\bottomrule
\end{tabular}
\caption{Performances of LLMs fine-tuned on atomic operation data on ARAOC and the 100 ARC tasks. %\lemao{Reorg this table to show the performance gap between ARROC and ARC. Merge the finetuning results together.}
}
\label{tab:fine-tune arc performance}
\end{table}
\fi

\section{Challenge on Task Composition}
\label{sec:factor}
In this section, we assess the deficiency of LLM's fluid intelligence from a task composition perspective. %To be specific, we evaluate the task composition ability of LLMs from two aspects. 
First, we consider a simple composition experiment that controllably evaluates the composition for Move and Copy in ARAOC (\sref{sec:simple composition}). Moreover, we design a complex composition experiment utilizing ARC tasks to evaluate LLMs' abilities to compose all atomic operations (\sref{sec:complex composition}).

\subsection{Evaluation on Simple Composition}
\label{sec:simple composition}
%One possible reason why LLMs fine-tuned on atomic operation data still fail on ARC is that their abilities to compose multiple atomic operations are weak. 
We start from evaluating LLMs' compositional ability on a simple composition task. 
To be specific, we compose Move and Copy to create 100 new tasks for evaluation. Since Mistral and Llama-3 are facing severe challenges on inducting these two atomic operations, we fine-tune them on three types of data: 1) 3,000 Move tasks; 2) 3,000 Copy tasks ; 3) 1,500 Move tasks and 1,500 Copy tasks, while making sure that these tasks do not overlap with those in ARAOC. We evaluate these fine-tuned LLMs and the GPT models on the newly crafted Move and Copy tasks and list the results in~\tref{tab:composition}.

%To further investigate the above assumption, 
%we fine-tune Mistral and Llama-3 on 3000 Move and Copy tasks that do not overlap with ARAOC (ensuring the number of fine-tuning examples remains the same as~\tref{tab:fine-tune arc performance}), respectively. Then, we test these fine-tuned LLMs on both the Move and Copy tasks in ARAOC, as well as on 100 additional tasks composed of Move and Copy operations. 

As can be seen, fine-tuning on single atomic operation's data can boost LLMs' performances on corresponding tasks, while fine-tuning on both atomic operations can achieve enhancement on both tasks. However, all the fine-tuned LLMs as well as GPT models face severe challenges when dealing with the composition tasks, which is not a complex composition, indicating that the composition abilities of LLMs are limited.

%Nevertheless, fine-tuning on both atomic operations' data could lead to better performances on the compositional task, raising up a straightforward question: \textit{could fine-tune with all the atomic operations help LLMs handle complex compositions in general ARC tasks?}

%This experimental result demonstrates our previous assumption that LLMs lack the capability to compose atomic operations, thus leads to their poor performances on ARC.
%.\mo{Conclusion1: Failure of transfer and composition of SFT paradigm}

%\lemao{This section is too short, you should consider how to add some new experiments or some contents? For example, you can use gpt4 for experiments without finetuning on both table 4 \& 5. In addition, you can add composition for other operations such as Change Color and Fill without finetuning?}

\subsection{Evaluation on Complex Composition}
\label{sec:complex composition}
%\lemao{Reorganize this subsection. Note that ARC can be considered as the compositions of different atomic operations.}

%Given that LLMs still perform poorly on ARAOC, 
%Similar to Section \ref{sec:evaluate on original arc}, we fine-tune Mistral and Llama-3 on tasks built upon atomic operations to see if this could enhance their performances on ARAOC. 
Furthermore, we examine the LLMs' abilities to compose atomic operations in complex ways. As mentioned in~\sref{sec: atom operation}, the ARC tasks can be decomposed into atomic operations listed in Table \ref{tab:atom operations}. Therefore, we regard ARC tasks as complex compositions of atomic operations for evaluation. Here we evaluate Llama-3 and GPT-4o since they are the better open-sourced and close-sourced LLMs in~\tref{tab:composition}.
In addition, % to the LLMs in~\sref{evaluated llms}, 
we fine-tune Llama-3 on tasks built upon atomic operations (check fine-tune details in Appendix~\ref{appendix:lora}) to see if this leads to improvement on ARC (\textbf{FT-atomic}).
%, using the same strategy and configurations as described in Section \ref{sec:evaluate on original arc}. Specifically, we generated an additional 500 tasks for each atomic operation beyond the 100 tasks in ARAOC, resulting in a total of 3000 tasks for fine-tuning. %We fine-tuned Mistral and Llama-3 on these data using the same strategy and configurations as described in Section \ref{sec:evaluate on original arc} (FT-atomic). 
In addition, we apply three more strategies to fine-tune Llama-3 for comparison: 1) using both the aforementioned operation data and 400 ARC tasks that do not overlap with the 100 evaluation tasks (\textbf{FT-atomic-arc}); 2) using only the 400 ARC tasks (\textbf{FT-arc}).
  %We evaluated these LLMs on ARC, and also listed their results on ARAOC for extra comparison.

Results are show in~\tref{tab:fine-tune arc performance}. We observe that fine-tuning on atomic operation data largely improves the performance of Llama-3 on ARAOC~\footnote{\scriptsize{We perform an additional experiment in Appendix~\ref{appendix:further fine-tuning} to further support this point.
}}. In particular, both fine-tuned LLMs achieve high accuracy on Color, Fill Internal, and Scale tasks, which Llama-3 struggles with. However, Llama-3-FT-atomic performs even worse than Llama-3 on ARC tasks. This could be due to the loss of compositional ability after solely fine-tuning on atomic operations, an issue that Llama-3-FT-atomic-arc does not encounter. On the other hand, fine-tuning on ARC tasks enhances LLMs' performance on ARC, but the improvement on ARAOC tasks is relatively limited compared to fine-tuning on ARAOC tasks. This is likely because the transformation rules in ARC are highly complex, and LLMs struggle to decompose these rules into atomic operations. Nonetheless, all LLMs still perform poorly on ARC tasks, which is unsurprising given their difficulties with even the simple compositions presented in~\tref{tab:composition}.
 
 %Overall, the experiment outcome again suggests that even if LLMs have a good understanding of individual atomic operations, they have limited ability to compose these atomic operations, causing them to fail on complex inductive reasoning tasks.

\textbf{Overall, while fine-tuning on atomic operations may assist LLMs in understanding these operations, it does not enable them to infer such operations from in-context examples. This limitation explains LLMs' poor performance on compositional tasks and further highlights their lack of intrinsic mechanisms for abstract reasoning, a core characteristic of fluid intelligence.
}

%the reason LLMs struggle with ARC tasks may be due to their weak ability to compose different atomic operations into the complex transformation rules required for ARC tasks.
 %\mo{Composition is one of the important aspects. The problems left are: (1) the LLMs learned the six operations instead of induction (this is actually the problem of 1DARC design); (2) the six operations do not cover every operation in ARC.}

%% file: sections/input.tex
\begin{table}[tb]
\small
\centering
\setlength{\tabcolsep}{1.3mm}
\begin{tabular}{ll|cccc}
\toprule
& \textbf{Setting} & \textbf{Mistral} & \textbf{Llama-3} & \textbf{GPT-3.5} & \textbf{GPT-4o} \\
\midrule[0.5pt]
\multirow{2}{*}{\textbf{Move}}& Ori & 2.00 & 1.00 & 4.00 & 13.00 \\
& Small & 12.00& 12.00 & 20.00 & 28.00\\
\midrule
\multirow{2}{*}{\textbf{Copy}}&Ori & 2.00 & 4.00 & 4.00 & 15.00 \\
&Small &12.00 &9.00 &14.00 & 34.00 \\
\bottomrule
\end{tabular}
\caption{Acc (in percentage) of LLMs with different input sizes. See~\tref{tab:large matrix_plus} for the Not M\% scores.}
\vspace{-0.1in}
\label{tab:large matrix}
\end{table}

\iffalse
\begin{table}[tb]
\small
\centering
\setlength{\tabcolsep}{1mm}
\begin{tabular}{ll|cccccc}
\toprule
& \multirow{2}{*}{\textbf{Setting}} & \multicolumn{2}{c}{\textbf{Mistral}} & \multicolumn{2}{c}{\textbf{Llama-3}} & \multicolumn{2}{c}{\textbf{GPT}} \\
\cmidrule(lr){3-4} \cmidrule(lr){5-6} \cmidrule(lr){7-8}
&~ & \textbf{7B} & \textbf{8*7B} & \textbf{8B} & \textbf{70B} &\textbf{3.5} &\textbf{4o}\\
\midrule[0.5pt]
\multirow{2}{*}{\textbf{Move}} & Ori & 2.00 && 1.00 && 4.00 & 13.00 \\
& Small & 12.00 & & 12.00 && 20.00 & 28.00 \\
\midrule
\multirow{2}{*}{\textbf{Copy}} & Ori & 2.00 && 4.00 && 4.00 & 15.00 \\
& Small & 12.00 && 9.00 && 14.00 & 34.00 \\
\bottomrule
\end{tabular}
\caption{Acc (in percentage) of LLMs with different input sizes. See~\tref{tab:large matrix_plus} for the Not M\% scores.}
\vspace{-0.1in}
\label{tab:large matrix}
\end{table}
\fi

\begin{table}[tb]
\small
\centering
\setlength{\tabcolsep}{2mm}
\begin{tabular}{lcc|cc}
\toprule
\textbf{LLM} &\textbf{Move} & \textbf{Copy} & \textbf{Comp} %& \textbf{Rank}
\\
\midrule
$\text{Mistral-FT}_{\text{Move}}$ & 19.00 & 4.00& 0.00  \\
$\text{Mistral-FT}_{\text{Copy}}$ &11.00 &32.00 & 2.00\\
$\text{Mistral-FT}_{\text{Move+Copy}}$ & 25.00& 32.00&4.00\\
\midrule
$\text{Llama-3-FT}_{\text{Move}}$ &21.00 &4.00 &0.00\\
$\text{Llama-3-FT}_{\text{Copy}}$ &12.00 &33.00 & 3.00\\
$\text{Llama-3-FT}_{\text{Move+Copy}}$ &26.00 & 27.00&5.00\\
\midrule
GPT-3.5 &4.00 &4.00 & 0.00\\
%GPT-4 &14.00 &13.00 & 4.00\\
GPT-4o &13.00 &15.00 & 2.00\\
\bottomrule
\end{tabular}
\caption{Acc (in percentage) on tasks composing Move and Copy (Comp).
%fine-tuned on single and multiple tasks. 
%Comp refers to tasks composed of Move and Copy. 
See~\tref{tab:composition_plus} for the Not M\% scores.}
\vspace{-0.2in}
\label{tab:composition}
\end{table}

\begin{table*}[tb]
\renewcommand\arraystretch{1.1}
\centering
\setlength{\tabcolsep}{2.5mm}
\small
\begin{tabular}{lcccccc|c}
\toprule[1pt]
\multirow{2}*{LLM} & \multicolumn{6}{c}{\textbf{Individual Atomic Operation}} & \multicolumn{1}{c}{\textbf{ Composition}} \\
\cmidrule{2-8}
& \multicolumn{1}{c}{\textbf{Move}} & \multicolumn{1}{c}{\textbf{Change Color}} & \multicolumn{1}{c}{\textbf{Copy}} & \multicolumn{1}{c}{\textbf{Mirror}} & \multicolumn{1}{c}{\textbf{Fill Internal}} & \multicolumn{1}{c}{\textbf{Scale}} & \multicolumn{1}{c}{\textbf{ARC}}  \\
% & Acc$\uparrow$ & Acc$\uparrow$ & Acc$\uparrow$ & Acc$\uparrow$ & Acc$\uparrow$ & Acc$\uparrow$ & Acc$\uparrow$ \\
\midrule[0.5pt]
%Mistral & 2.00 & 15.00 & 2.00 & 1.00 & 9.00 & 0.00 & 2.00 \\
%$\text{Mistral-FT-atomic}$ & 12.00 & 100.00 & 20.00 & 26.00 & 97.00 & 89.00 & 1.00 \\
%$\text{Mistral-FT-atomic-arc}$ &14.00 &99.00&14.00&24.00&97.00&87.00&6.00\\
%\midrule
Llama-3 & 1.00 & 39.00 & 4.00 & 2.00 & 63.00 & 1.00 & 5.00 \\
$\text{Llama-3-FT-arc}$ &2.00 &73.00 &5.00 &3.00 &88.00 &0.00 &9.00 \\
$\text{Llama-3-FT-atomic}$ & 13.00 & 98.00 & 14.00 & 27.00 & 97.00 & 78.00 & 2.00 \\
$\text{Llama-3-FT-atomic-arc}$ & 12.00 &97.00&17.00&28.00&98.00&79.00&6.00 \\
\midrule
%GPT-3.5 &4.00&48.00&4.00&6.00&58.00&1.00 &6.00 \\
%GPT-4 &14.00 &97.00&13.00&14.00&100.00&3.00&17.00 \\
GPT-4o & 13.00&98.00&15.00&12.00&96.00&2.00&19.00 \\
\bottomrule[1pt]
\end{tabular}
\caption{Results of LLMs on individual and composition of atomic operations. %FT refers to fine-tune on the atomic operation data. S
See~\tref{tab:fine-tune arc performance_plus} for the Not M\% scores.}
\vspace{-0.2in}
\label{tab:fine-tune arc performance}
\end{table*}

\begin{table}[tb]
\small
\centering
\setlength{\tabcolsep}{3mm}
\begin{tabular}{lcccc}
\toprule
\textbf{LLM} &\textbf{Size} & \textbf{Location} & \textbf{Transpose} %& \textbf{Rank}
\\
\midrule
Mistral &0.32 &0.00 &0.02 %&0.29 
\\
Llama-3 & 0.63&0.04 &0.04 %&0.86
\\
%Mistral 8*7B & & & %&0.29 
%\\
%Llama-3 70B & & & %&0.86
%\\
\midrule
GPT-3.5 &0.93 &0.43 &0.34 %&0.30
\\
%GPT-4o &\textbf{1.00} &\textbf{0.77} &\textbf{0.83} \\%s&0.95\\
GPT-4o &\textbf{1.00} &\textbf{0.91} &\textbf{0.91} \\
\bottomrule
\end{tabular}
\caption{LLMs' accuracy on matrix-related questions. The best results under each column are \textbf{boldfaced}.}
\vspace{-0.1in}
\label{tab:understand matrix}
\end{table}

\section{Challenge on Input Format}
\label{sec:matrix}

Since LLMs cannot process visual inputs, we follow~\citet{wang2023hypothesis} to convert the 2D visual input-output grids in ARAOC tasks into matrix-format before feeding them to the LLMs (\sref{sec:arc setting}). However, it remains uncertain that whether this conversion affects LLMs' performances on ARAOC, since LLMs are mostly trained on natural language data, and may not understand such matrix-format inputs well. In this section, we first try to answer this question (\sref{sec:understand matrix}), then investigate a strategy to remedy its potential challenges (\sref{sec:natural language}).

\subsection{Matrix-format Understanding}
\label{sec:understand matrix}
%We analyze this problem from two perspectives. First, 
We first investigate whether LLMs understand the input matrices well. Specifically, we select the testing input matrices from the 100 ARAOC Move tasks, and ask LLMs to output the size, transpose, and subgrid's corner elements' locations of each matrix (see the input prompt in~\fref{fig:matrix property prompt}). Our intuition is that if LLMs correctly answer these questions, they should have understood the matrix-format input. % and this should not affect their performances on ARAOC tasks. 
Results are shown in~\tref{tab:understand matrix}, where GPT-4o answers these questions with high accuracy, indicating that it comprehends such matrix-format inputs well. However, other LLMs perform poorly on these tasks, which may further affect their results on ARAOC.

To further investigate the impact of matrix-format input, we re-evaluate $\text{Llama-3-FT}_{\text{Move+Copy}}$ from~\tref{tab:composition} and GPT-4o on the Move and Copy tasks without using the location information of subgrids, as detailed in Appendix~\ref{appendix:banning}. The results in Appendix~\ref{appendix:banning} show that prohibiting the use of location information do reduce LLMs' performances on both tasks, indicating that a fundamental understanding of matrices is crucial for completing ARAOC and ARC tasks. However, as the combined results from~\tref{tab:araoc results} and~\tref{tab:understand matrix} suggest, possessing matrix understanding alone does not guarantee good performance on these tasks.

\begin{table}[tb]
\small
\centering
\setlength{\tabcolsep}{0.5mm}
\begin{tabular}{ll|cccc}
\toprule
& \textbf{Method}& \textbf{Mistral}& \textbf{Llama-3} & \textbf{GPT-3.5}%& \textbf{GPT-4}
&\textbf{GPT-4o}\\
\midrule[0.5pt]
\multirow{2}{*}{\textbf{Move}}& w/o NL &2.00 &1.00 &4.00 & 13.00\\%&14.00 \\
& NL & 5.00 &12.00 &23.00 &53.00 \\%49.00 \\
\midrule[0.5pt]
\multirow{2}{*}{\textbf{Color}}& w/o NL &15.00 &39.00 &48.00 & 98.00\\%97.00 \\
& NL &3.00 &83.00 &59.00 &99.00 \\%92.00 \\
\midrule[0.5pt]
\multirow{2}{*}{\textbf{Copy}}& w/o NL &2.00 &4.00 &4.00 & 15.00\\%13.00 \\
& NL &2.00 &6.00 &14.00 &40.00 \\%45.00 \\
\midrule[0.5pt]
\multirow{2}{*}{\textbf{Mirror}}& w/o NL &1.00 &2.00 &6.00 & 12.00\\%14.00 \\
& NL &2.00 &8.00 &21.00 & 30.00\\%29.00 \\
\midrule[0.5pt]
\multirow{2}{*}{\textbf{Fill Internal}}& w/o NL &9.00 &63.00 &58.00 &96.00 \\%100.00 \\
& NL &0.00 &10.00 &35.00 & 72.00\\%85.00 \\
\midrule[0.5pt]
\multirow{2}{*}{\textbf{Scale}}& w/o NL &0.00 &1.00 &1.00 &2.00 \\%3.00 \\
& NL &0.00 &2.00 &0.00 &4.00 \\%6.00 \\
\bottomrule
\end{tabular}
\caption{Acc (in percentage) of LLMs with natural language inputs (NL). %w/o NL refers to the results in~\tref{tab:araoc results}. 
See Not M \% scores in~\tref{tab:natural language input_plus}.}
\vspace{-0.2in}
\label{tab:natural language input}
\end{table}

\subsection{Switching Matrix into Natural Language}
\label{sec:natural language}
%Finally, we propose an approach to relieve the challenge on input understanding, which is inspired by our previous finding that GPT-4 struggles to derive correct transformation rules from ARAOC's input matrices (\tref{tab:araoc results}) despite its excellent understanding of these matrices (\tref{tab:understand matrix}). 
%From previous sections we conclude that the challenges of ARAOC and ARC tasks is not due to LLMs' understanding of matrix-format inputs.
%Therefore, a new type of inputs for representing such tasks is necessary. 
Since LLMs are predominantly trained on natural language rather than matrix-format data, we further propose to convert the matrix-format input-output grids into natural language with the aid of a coordinate system-based prompt (listed in~\fref{fig:natural language prompt}). We evaluate LLMs using this new prompt on ARAOC, and the results are presented in~\tref{tab:natural language input}.

Notably, we find that on tasks that LLMs originally cannot answer well (Move, Copy, Mirror, and Scale), using natural language inputs can largely boost their performances. As for tasks that are relatively easy for LLMs, converting matrix-format input to natural language still keep the good performances. %These results illustrate the effectiveness of our proposed approach. 
One exception appears to be the Mistral model, whose performance decreases with the natural language prompt. This is probably because this model is not strong enough to encode the natural language input that can be handled by other LLMs, which makes its results not indicative.

\textbf{Overall, we conclude that LLMs' failure on fluid intelligence tests is not mainly due to their understanding of the specific matrix-format inputs, but their limitations on encoding such inputs for obtaining global representations of the input tasks.}

%% file: sections/model.tex
\section{Modeling Challenge}
\label{sec:model}

%As LLMs have unique model architectures and text encoding strategies, we are also interested in whether these modeling features affect their performances on ARAOC. 
In this section, we examine whether LLMs' modeling features affect their fluid intelligence from both the model architecture perspective and the information encoding perspective.

\iffalse
\begin{table}[tb]
\small
\centering
\setlength{\tabcolsep}{1mm}
\begin{tabular}{l|cccccc}
\toprule
\multirow{2}{*}{\textbf{Direction}} & \multicolumn{2}{c}{\textbf{Mistral}} & \multicolumn{2}{c}{\textbf{Llama-3}} & \textbf{GPT-3.5} & \textbf{GPT-4o} \\
\cmidrule(lr){2-3} \cmidrule(lr){4-5} %\cmidrule(lr){6-7}
~ & \textbf{7B} & \textbf{8*7B} & \textbf{8B} & \textbf{70B} & &\\
\midrule[0.5pt]
Left &4.00 && 2.00&11.00 &11.00 & 13.00\\

Right &4.00& &6.00&17.00 &20.00** & 28.00**\\

\bottomrule
\end{tabular}
\caption{Acc (in percentage) of LLMs with two mirroring directions. ``**'' means 
 the bottom result is significant better than the upper one with p < 0.05. See~\tref{tab:autoregressive_plus} for the Not M\% scores.}
\vspace{-0.2in}
\label{tab:autoregressive}
\end{table}
\fi
\subsection{The Bias of Model Architecture}
\label{sec:model architecture}
%\paragraph{Does the Autoregressive Characteristic of LLMs Affects Their Performances?}

When predicting output tokens given an input prompt, existing LLMs use the autoregressive decoding strategy~\cite{bahdanau2014neural}, which predicts the next token based solely on previous tokens. However, in some ARAOC tasks like Mirror, the newly generated part in the testing output grid may locate before the original part. This prevents LLMs from using information in the testing input grid to generate the new part, thus lowering their performances. For example, if the Mirror example in~\tref{tab:atom operations} is a testing input-output grid pair, LLMs cannot reference the bottom two green grids (the original subgrid) while generating the upper two green grids (the new subgrid), which makes the generation process more challenging.

To investigate this hypothesis, we conduct an experiment using the Mirror operation. Specifically, 
we first randomly generate 100 new input grids for Mirror, while lowering the number of rows and columns of input grids within [3, 7] to get more significant results. 
We then mirror each input grid towards left and right to create two individual tasks and overall leads to 100 tasks for left and right, respectively. We evaluate all the LLMs on these tasks and perform a binomial significance test to examine the differences in their performance across both directions. %We test all the LLMs on these tasks and present the 

Results are shown in~\tref{tab:autoregressive}. Noting that stronger models (the GPT models) perform significantly better when the mirroring direction is to the right, i.e., when the original subgrid is predicted before the mirrored one. This supports our hypothesis that the autoregressive nature of LLMs hinders their performance, as it prevents the simultaneous back and forth processing required by fluid intelligence. For weaker models, their relatively low scores render their results less conclusive, although Llama-3 still achieves higher Acc scores when the mirroring direction is to the right, which aligns with our hypothesis. Additionally, to further explore whether the above findings hold for LLMs of different sizes, we evaluate Mistral-8*7B and Llama-3-70B on these tasks and provide a detailed analysis in Appendix~\ref{appendix:additional table12}.

%\mo{Conclusion: autoregressive decoding}

\begin{table}[tb]
\small
\centering
\setlength{\tabcolsep}{2mm}
\begin{tabular}{l|llll}
\toprule
\textbf{Direction}& \textbf{Mistral}& \textbf{Llama-3} & \textbf{GPT-3.5}& \textbf{GPT-4o} \\
\midrule[0.5pt]
Left &4.00 & 2.00 &11.00 & 13.00\\

Right &4.00 &6.00 &20.00** & 28.00**\\

\bottomrule
\end{tabular}

\caption{Acc (in percentage) of LLMs with two mirroring directions. ``**'' means 
 the bottom result is significant better than the upper one with p < 0.05. See~\tref{tab:autoregressive_plus} for the Not M\% scores.}
\vspace{-0.2in}
\label{tab:autoregressive}
\end{table}

\subsection{Challenge on Information Usage}
\label{sec:information usage}
Since each task in ARAOC includes 3 in-context examples, %where the transformation rule is learned from several input-output grids and then applied to predict a testing output grid, 
the ability of LLMs to identify useful information from the in-context examples may affect their performances on ARAOC. We investigate this claim by calculating the saliency score~\cite{simonyan2013deep} of one of Mistral's incorrect predictions with respect to the in-context examples, %as saliency scores highlight which parts of the input most affect the prediction, 
with higher scores indicating a larger impact. %Specifically, we select a task representing the Move operation that Mistral fails to complete in~\tref{tab:large matrix}. Then, we calculate and visualize the saliency scores of tokens in the task's input prompt with respect to the number "6" that Mistral is supposed to move, using the LIT tool~\cite{tenney2020language}.

\begin{figure}
    \centering
    \includegraphics[width=0.48\textwidth]{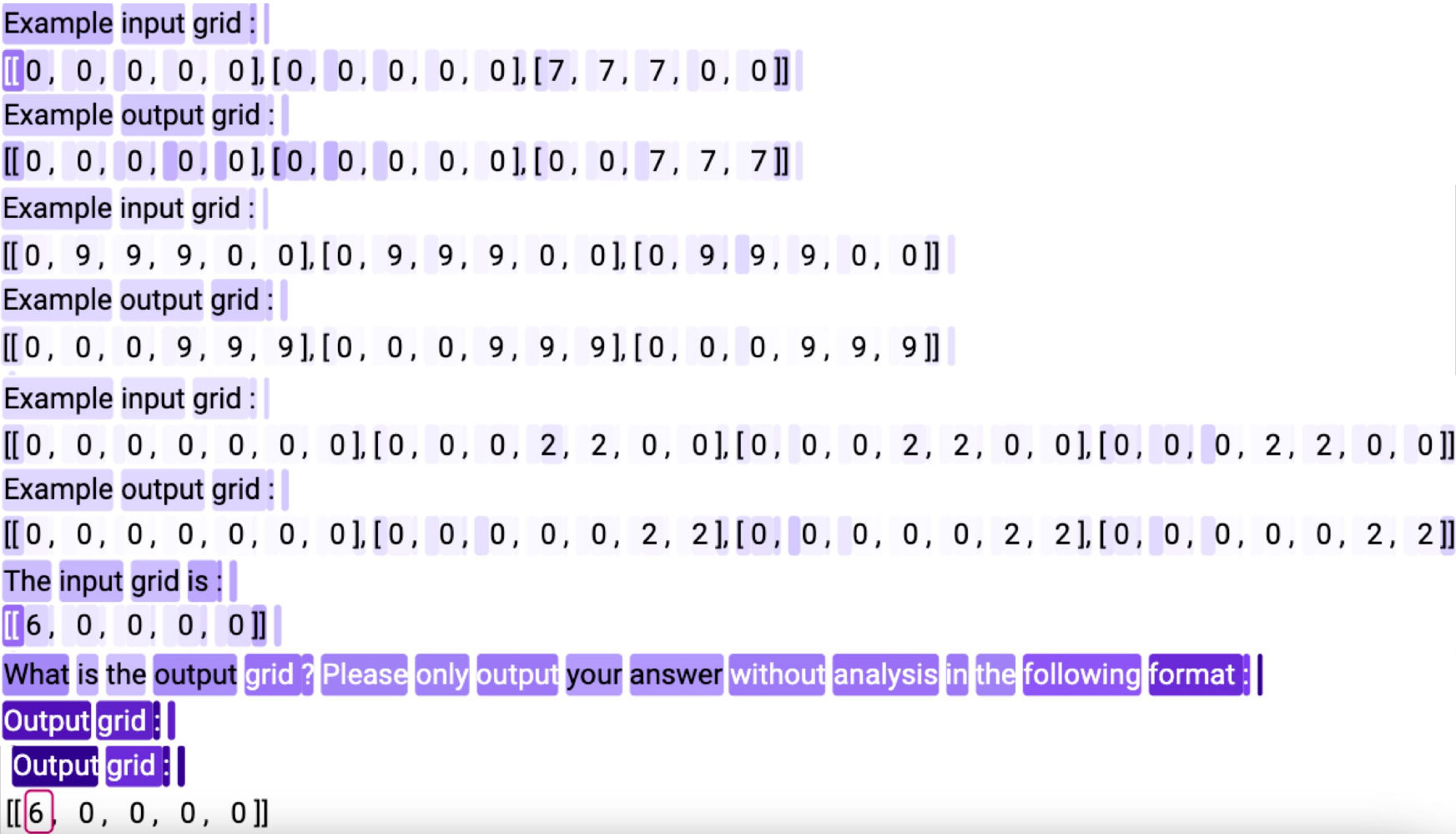}
    \caption{A saliency analysis example, where darker means higher saliency corresponds to the boxed token.
} 
\vspace{-0.2in}
    \label{fig:saliency}
\end{figure}

As shown in~\fref{fig:saliency}, Mistral should move "6" two steps to the right, %, as illustrated by the in-context examples. 
yet it incorrectly keeps "6" fixed in the output grid. % leading to the prediction error. 
With the saliency scores, we find that Mistral does not focus much on the moved parts in the in-context examples (e.g., all the "7"s in the first example). Instead, it focuses more on %the title and 
the unchanged parts, which leads it mistakenly assume that "6" should also be fixed. These observations illustrate that the inability to identify relevant information in in-context examples also explains why LLMs struggle with ARAOC tasks. In addition, we provide a saliency analysis example where Mistral makes a correct prediction in Appendix~\ref{appendix:additional saliency} for comparison.

\textbf{Overall, we conclude that LLMs' %inherently lack fluid intelligence because their
internal architecture also
%(the autoregressive scanning direction (\sref{sec:model architecture}) and the (triangular) causal attention mask (\sref{sec:information usage})) 
limits their ability to access global information, which is important for illustrating fluid intelligence.
%LLM's autoregressive decoding characteristic makes them lack the global understanding for demostrating fluid intelligence, which requires both forth and back information.
} While the findings on LLMs' fluid intelligence in previous sections are drawn from ARC and ARAOC tasks, they can be generalized to other real-world tasks and a further discussion on the applicability of our findings can be found in Appendix~\ref{appendix:generalization}.

%Mistral tends to focus more on parts that do not contribute to the final predictions. For instance, when predicting the number 8 for the Move case, an ideal LLM should concentrate on the number 8 in the input grid. However, Mistral directs more attention to the number 0 following the 8, causing it to incorrectly predict the position of 8 in the output grid. These observations lead us to conclude that the inability to correctly identify relevant information in in-context examples is a significant factor in why LLMs struggle with ARAOC tasks.

%% file: sections/related_work.tex
\section{Related Works}

\paragraph{Evaluating fluid intelligence of LLMs.}
As an essential aspect of intelligence~\cite{cattell1963theory, cattell1971abilities, jaeggi2008improving}, studying the fluid intelligence of LLMs offers deeper insights into their overall intelligence. \citet{chollet2019measure} and \citet{barak2024investigating} suggest that abstract inductive reasoning is an ideal method for evaluating LLMs' fluid intelligence. However, most existing benchmarks~\cite{honovich2023instruction, yang2024language, qiuphenomenal} fail to prevent memorization shortcuts, making them easier for LLMs to solve. In contrast, the ARC corpus~\cite{chollet2019measure} that challenges models to identify transformation rules between input-output grids, poses significant difficulty for LLMs, making it suitable for fluid intelligence assessment. 
Previous works have primarily focused on improving LLM performance on ARC tasks~\cite{min2023approach, tan2023large, xullms, mirchandani2023large, wang2024speak, huang2024anpl, wang2023hypothesis}, but the results remain far from optimal. This motivates us to explore the underlying reasons behind LLMs' limited fluid intelligence.

\paragraph{Matrix operations with LLMs.}
It has been shown that LLMs have abilities to understand matrix operations% like transpose and inverse, 
%and fine-tuning on matrix operation data can enhance LLMs' performances on matrix-related problems
~\cite{charton2021linear, collins2024evaluating, azerbayevllemma, shao2024deepseekmath}. However, \sref{sec:understand matrix} indicates that understanding the properties of matrix may not the key factor of LLMs' success on ARC and ARAOC, leading to further analyses.

%% file: sections/appendix.tex
\section{Further details regarding the experiments in~\tref{tab:inductive reasoning examples}}
\label{appendix:inductive examples} 
In~\tref{tab:inductive reasoning examples}, we also measure GPT-4o and human's performances on the first three inductive reasoning tasks. The details are as follows:
\begin{enumerate}
    \item \textit{II} refers to \textit{Instruction Induction}. We conduct experiment on the ``Synonyms'' task in~\citet{honovich2023instruction}, since models in~\cite{honovich2023instruction} obtain the worst performance on this subtask. After GPT-4o has generated all the responses, instead of using BERT-Score to evaluate the responses, we ask a human annotator to give 0/1 (incorrect/correct) points to each response based on the ground truths. Then we calculate GPT-4o's average score as the final performance. Human's performance is extracted from~\citet{honovich2023instruction}.
    \item As for Deer, since it does not provide human performance, we invite an annotator to manually finish the first 50 tasks in its test set, and ask another annotator to give 0/1 (incorrect/correct) points to each human response based on the ground truths. We also ask GPT-4o to finish the first 50 tasks and do the same. Then we calculate GPT-4o and human's average score as the final performance.
    \item As for Mini Scan, the results are extracted from~\cite{qiuphenomenal}.
\end{enumerate}
\textit{II} refers to \textit{Instruction Induction}. We experiment on its ``Synonyms'' task, where models obtain the worst performance. As for Deer, we evaluate on the first 50 tasks of its test set

\section{The Prompts We Use in this Paper}
\label{appendix:prompts}
All the prompt templates we use in this paper are listed in~\fref{fig:original prompt}, \fref{fig:visual prompt}, \fref{fig:visual+textual prompt}, \fref{fig:visual+textual prompt1}, \fref{fig:matrix property prompt}, \fref{fig:without location prompt}, and \fref{fig:natural language prompt}.

\begin{figure*}
  \begin{tcolorbox}
  \textbf{SYSTEM}:\\
  You are a helpful assistant.\\\\
  \textbf{USER}: \\
        You will be playing a game that need to find common patterns from input examples and apply the pattern for prediction on new examples.\\
Lets play a game where you are transforming an input grid of numbers into an output grid of numbers.\\

The numbers represent different colors:\\
0 = black\\
1 = blue\\
2 = red\\
3 = green\\
4 = yellow\\
5 = gray\\
6 = magenta\\
7 = orange\\
8 = cyan\\
9 = brown\\

Here are examples of input grids and its corresponding output grids:\\
Example input grid:\\
\{INPUT GRID 1\} \\
Example output grid:\\
\{OUTPUT GRID 1\} \\

Example input grid:\\
\{INPUT GRID 2\} \\
Example output grid:\\
\{OUTPUT GRID 2\} \\

Example input grid:\\
\{INPUT GRID 3\} \\
Example output grid:\\
\{OUTPUT GRID 3\} \\

The input grid is:

\{TESTING INPUT GRID\} \\

What is the output grid? Please only output your answer without analysis in the following format:

Output grid:

    \end{tcolorbox}
    \caption{The standard prompt we use in this paper that converts ARC/ARAOC tasks into matrix-format inputs. Also the prompt for the textual input/textual output setting in~\tref{tab:different format}.}
    \label{fig:original prompt}
\end{figure*}

\begin{figure*}
  \begin{tcolorbox}
  \textbf{SYSTEM}:\\
  You are a helpful assistant.\\\\
  \textbf{USER}: \\
        You will be playing a game that need to find common patterns from input examples and apply the pattern for prediction on new examples.\\

\{IMAGE\} \\

In the given image, there are two columns of matrices with elements represented by different colors. The left column contains the input matrices, and the right column contains the corresponding output matrices. The last row includes only an input matrix, while the other rows include both input and output matrices. Your task is to identify the pattern from the given input-output matrix pairs and apply this pattern to predict the output matrix for the input matrix in the last row. \\

Please complete the task by generating an image that includes only the predicted output matrix.

    \end{tcolorbox}
    \caption{The prompt for the visual input/visual output setting in~\tref{tab:different format}.}
    \label{fig:visual prompt}
\end{figure*}

\begin{figure*}
  \begin{tcolorbox}
  \textbf{SYSTEM}:\\
  You are a helpful assistant.\\\\
  \textbf{USER}: \\
        You will be playing a game that need to find common patterns from input examples and apply the pattern for prediction on new examples.\\
Lets play a game where you are transforming an input grid of numbers into an output grid of numbers.\\

The numbers represent different colors:\\
0 = black\\
1 = blue\\
2 = red\\
3 = green\\
4 = yellow\\
5 = gray\\
6 = magenta\\
7 = orange\\
8 = cyan\\
9 = brown\\

Here are examples of input grids and its corresponding output grids:\\
Example input grid:\\
\{INPUT GRID 1\} \\
Example output grid:\\
\{OUTPUT GRID 1\} \\

Example input grid:\\
\{INPUT GRID 2\} \\
Example output grid:\\
\{OUTPUT GRID 2\} \\

Example input grid:\\
\{INPUT GRID 3\} \\
Example output grid:\\
\{OUTPUT GRID 3\} \\

\{IMAGE\} \\

The 2D format of these input and output grids are also provided in the given image for your reference. \\

The input grid is:

\{TESTING INPUT GRID\} \\

What is the output grid? Please generate an image of the output grid similar to those in the given image, do not output any text.

    \end{tcolorbox}
    \caption{The prompt for the visual+textual input/visual output setting in~\tref{tab:different format}.}
    \label{fig:visual+textual prompt}
\end{figure*}

\begin{figure*}
  \begin{tcolorbox}
  \textbf{SYSTEM}:\\
  You are a helpful assistant.\\\\
  \textbf{USER}: \\
        You will be playing a game that need to find common patterns from input examples and apply the pattern for prediction on new examples.\\
Lets play a game where you are transforming an input grid of numbers into an output grid of numbers.\\

The numbers represent different colors:\\
0 = black\\
1 = blue\\
2 = red\\
3 = green\\
4 = yellow\\
5 = gray\\
6 = magenta\\
7 = orange\\
8 = cyan\\
9 = brown\\

Here are examples of input grids and its corresponding output grids:\\
Example input grid:\\
\{INPUT GRID 1\} \\
Example output grid:\\
\{OUTPUT GRID 1\} \\

Example input grid:\\
\{INPUT GRID 2\} \\
Example output grid:\\
\{OUTPUT GRID 2\} \\

Example input grid:\\
\{INPUT GRID 3\} \\
Example output grid:\\
\{OUTPUT GRID 3\} \\

\{IMAGE\} \\

The 2D format of these input and output grids are also provided in the given image for your reference. \\

The input grid is:

\{TESTING INPUT GRID\} \\

What is the output grid? Please only output your answer without analysis in the following format:

Output grid:

    \end{tcolorbox}
    \caption{The prompt for the visual+textual input/textual output setting in~\tref{tab:different format}.}
    \label{fig:visual+textual prompt1}
\end{figure*}

\begin{figure*}
  \begin{tcolorbox}
  \textbf{SYSTEM}:\\
  You are a helpful assistant.\\\\
  \textbf{USER}: \\
Given a matrix in the format of numpy array, please answer the following questions:

1. What is the size of this matrix?  Output in the format of (a,b).

2. What is the location of the non-zero subgrids. Please first find out all the corner elements of the subgrids, then output their locations in the order of [top-left, top-right, bottom-left, bottom-right], in the format of (which row, which col).

3. What is the transpose of this matrix? Output the transposed matrix in the format of a numpy array with elements separated by commas and enclosed in square brackets for each row like "[[0, 0, 0, 0], [0, 0, 0, 0], [0, 0, 0, 0], [0, 0, 0, 0]]".

4. What is the rank of this matrix? Output the rank of the matrix.\\

Please only output your answer without analysis in the following format:

1.Size: 

2.Location: 

3.Transpose:

4.Rank:\\

Input Matrix: 

\{INPUT MATRIX\}

    \end{tcolorbox}
    \caption{The prompt for asking LLMs about matrix properties, which is used in~\sref{sec:understand matrix}.} 
    \label{fig:matrix property prompt}
\end{figure*}

\begin{figure*}
  \begin{tcolorbox}
  \textbf{SYSTEM}:\\
  You are a helpful assistant.\\\\
  \textbf{USER}: \\
        You will be playing a game that need to find common patterns from input examples and apply the pattern for prediction on new examples.\\
Lets play a game where you are transforming an input grid of numbers into an output grid of numbers.\\

The numbers represent different colors:\\
0 = black\\
1 = blue\\
2 = red\\
3 = green\\
4 = yellow\\
5 = gray\\
6 = magenta\\
7 = orange\\
8 = cyan\\
9 = brown\\

Here are examples of input grids and its corresponding output grids:\\
Example input grid:\\
\{INPUT GRID 1\} \\
Example output grid:\\
\{OUTPUT GRID 1\} \\

Example input grid:\\
\{INPUT GRID 2\} \\
Example output grid:\\
\{OUTPUT GRID 2\} \\

Example input grid:\\
\{INPUT GRID 3\} \\
Example output grid:\\
\{OUTPUT GRID 3\} \\

The input grid is:

\{TESTING INPUT GRID\} \\

What is the output grid? When answering this question, please avoid using information about: 1) the sizes of the input grids and the output grids; 2) the locations of different numbers in the input grids and the output grids. \\

Please only output your answer without analysis in the following format:

Output grid:

    \end{tcolorbox}
    \caption{Prompt that bans the use of location information.}
    \label{fig:without location prompt}
\end{figure*}

\begin{figure*}
  \begin{tcolorbox}
  \textbf{SYSTEM}:\\
  You are a helpful assistant.\\\\
  \textbf{USER}: \\
        You will be playing a game that need to find common patterns from input examples and apply the pattern for prediction on new examples.\\
Lets play a game where you are transforming an input grid of numbers into an output grid of numbers.\\

The numbers represent different colors:\\
0 = black\\
1 = blue\\
2 = red\\
3 = green\\
4 = yellow\\
5 = gray\\
6 = magenta\\
7 = orange\\
8 = cyan\\
9 = brown\\

Here are examples of input grids and its corresponding output grids:\\
Example input grid:\\
The matrix dimensions are \{\} columns by \{\} rows. Coordinates are based on a Cartesian coordinate system with the origin (0,0) at the bottom-left corner. The coordinates of the non-zero elements, listed from top to bottom and left to right, are: \{\} \\
Example output grid:\\
The matrix dimensions are \{\} columns by \{\} rows. Coordinates are based on a Cartesian coordinate system with the origin (0,0) at the bottom-left corner. The coordinates of the non-zero elements, listed from top to bottom and left to right, are: \{\}\\

...... (leave out input-output grid pairs 2 and 3) \\

The input grid is:

The matrix dimensions are \{\} columns by \{\} rows. Coordinates are based on a Cartesian coordinate system with the origin (0,0) at the bottom-left corner. The coordinates of the non-zero elements, listed from top to bottom and left to right, are: \{\} \\

What is the output grid?  \\

Please only output your answer without analysis in the following format:

Output grid:

    \end{tcolorbox}
    \caption{Prompt that converts matrix-format input to natural language.}
    \label{fig:natural language prompt}
\end{figure*}

\section{Detailed analysis regarding the failure of visual-based LLMs on ARC}
\label{appendix:visual analysis}
In~\tref{tab:different format} we find that it is extremely for visual-based LLMs to finish ARC tasks. After manually checking the model responses, we conlcude that it is because when answering ARC tasks, the visual-based LLMs needs to generate every small pixel (grid) correctly to form a totally correct output grid, which is extremely challenging for visual-based LLMs like GPT-4o. 

To take a deeper look at how visual-based GPT-4o fails on ARC, we sample a few grids from ARC instances where each grid has a size $\leq 10\times10$. We then take steps to ask GPT-4o to recognize the grid from the image and convert it to the textual matrix format. GPT-4o manages to recognize the sizes with around 50\% accuracy (considering each instance consists of more than 6 grids, this would result in large error propagation). Additionally, GPT-4o fails to correctly convert any grid to the matrix format. This study further illustrates that GPT-4o lacks the ability to ground the figures of grids to the symbolic space, consequently limiting its reasoning performance.

\section{Inference Configurations of LLMs}
\label{appendix:inference config}
For GPT models, we use their default inference configurations mentioned in \url{https://platform.openai.com/docs/guides/text-generation/completions-api}. As for Mistral and Llama, we set the maximum output length to be 3000 tokens, and follow their default settings for other configurations. During inference, for all the models, we maintain their official prompt templates unchanged.

\section{Fine-tuning Details}
\label{appendix:lora}
For all the fine-tuning experiments, we do not fine-tune all the LLM's parameters, and use LoRA instead, as mentioned in~\sref{sec:evaluate on original arc}. We fine-tune each model for 3 epochs with a batch size of 4. The dimension of LoRA's attention layer is set to 64, and the $\alpha$ and dropout rates are set to 16 and 0.1, respectively. The learning rate and weight decay are set to 2e-4 and 0.001, respectively.

For the fine-tuning data used in~\sref{sec:complex composition}, we generated an additional 500 tasks for each atomic operation beyond the 100 tasks in ARAOC, resulting in a total of 3000 tasks for the FT-atomic fine-tuning.

\section{Analysis on whether LLMs Learn Atomic Operations During Fine-tuning}
\label{appendix:further fine-tuning}

\begin{table*}[tb]
\renewcommand\arraystretch{1.1}
\centering
\setlength{\tabcolsep}{1.7mm}
\small
\begin{tabular}{lcccccc}
\toprule[1pt]
\multirow{2}*{LLM} & \multicolumn{6}{c}{\textbf{Individual Atomic Operation}} \\
\cmidrule{2-7}
& \multicolumn{1}{c}{\textbf{Move}} & \multicolumn{1}{c}{\textbf{Change Color}} & \multicolumn{1}{c}{\textbf{Copy}} & \multicolumn{1}{c}{\textbf{Mirror}} & \multicolumn{1}{c}{\textbf{Fill Internal}} & \multicolumn{1}{c}{\textbf{Scale}} \\

\midrule[0.5pt]

Llama-3 & 1.00 & 39.00 & 4.00 & 2.00 & 63.00 & 1.00   \\
$\text{Llama-3-FT-atomic}$ & 13.00 & 98.00 & 14.00 & 27.00 & 97.00 & 78.00  \\
$\text{Llama-3-FT-atomic w/o own}$ &10.00 (17.00)&94.00 (4.00)&6.00 (22.00)&5.00 (81.00)&58.00 (0.00)&2.00 (96.00) \\

\bottomrule[1pt]
\end{tabular}
\caption{Results of LLMs fine-tuned on different atomic operations. Not M\% scores of Llama-3-FT-atomic w/o own are shown in brackets. Not M\% scores for other models are listed in~\tref{tab:fine-tune arc performance_plus}.}
\vspace{-0.2in}
\label{tab:fine-tune different}
\end{table*}

In~\tref{tab:fine-tune arc performance}, we observe that fine-tuning on atomic operations enhances LLM performance on ARAOC tasks. However, it is possible that these improvements come from the LLMs learning the new matrix format of the input/output, rather than truly learning the atomic operations. To further investigate this, we conducted additional experiments. For Llama-3-FT-atomic in~\tref{tab:fine-tune arc performance}, we fine-tuned Llama-3 on all six atomic operations, using 500 tasks for each. In the new experiments, for each atomic operation, we fine-tuned Llama-3 on the other five atomic operations, using 600 tasks for each, ensuring the total number of fine-tuning examples remained consistent. The resulting model (Llama-3-FT-atomic w/o own) was then tested on the excluded atomic operation. The rationale behind this setup is that if the performance improvement observed in Llama-3-FT-atomic was solely due to the model learning the new matrix format of the input and output, rather than the atomic operations, the performance of Llama-3-FT-atomic w/o own should be similar to Llama-3-FT-atomic.

The results are presented in~\tref{tab:fine-tune different}. As shown, while Llama-3-FT-atomic w/o own improves upon Llama-3, performance gaps remain between Llama-3-FT-atomic w/o own and Llama-3-FT-atomic. Based on these results and the analysis in~\tref{tab:fine-tune different}, we conclude that, to a large extent, the fine-tuning process enhances the original LLM's understanding of atomic operations.

\section{Details on Crafting ARAOC}
\label{appendix:araoc}
\begin{enumerate}
    \item \textbf{Move}: for the Move tasks, the numbers of rows and columns of the input and the output grids are randomly initialized from [1,16], where the numbers of rows and columns of the subgrid is randomly initialized from [1, min(a, b)+1]. The number of moving step is sampled from [1, 8]
       \item \textbf{Change Colour}: for the Change Color tasks, the numbers of rows and columns of the input and the output grids are randomly initialized from [1,16], where the numbers of rows and columns of the subgrid is randomly initialized from [1, min(a, b)+1]. The new color is randomly sampled from the ten colors, while not overlapping with the original color.
   \item \textbf{Copy}: for the Copy tasks, the numbers of rows and columns of the input and the output grids are randomly initialized from [1,16], where the numbers of rows and columns of the subgrid is randomly initialized from [1, min(a, b)+1]. The number of steps between the copied subgrid and the original subgrid is sampled from [1, 8].
   \item \textbf{Mirror}: for the Mirror tasks, the numbers of rows and columns of the input and the output grids are randomly initialized from [1,16], where the numbers of rows and columns of the subgrid is randomly initialized from [1, min(a, b)+1]. 
   \item \textbf{Fill Internal}: for the Fill Internal tasks, the numbers of rows and columns of the input and the output grids are randomly initialized from [3,16], where the numbers of rows and columns of the subgrid is randomly initialized from [1, min(a, b)+1]. The filled color is randomly sampled from the ten colors.
\item \textbf{Scale}: for the Scale tasks, the numbers of rows and columns of the input and the output grids are randomly initialized from [2,5], where the numbers of rows and columns of the subgrid is randomly initialized from [1, min(a, b)+1]. 
\end{enumerate}

\begin{table}[tb]
\small
\centering
\setlength{\tabcolsep}{4mm}
\begin{tabular}{l|ccc}
\toprule
LLM& Move & Copy \\
\midrule
%$\text{Mistral-FT}_{\text{Move+Copy}}$ &25.00 &32.00 \\
 %\quad \quad \quad w/o Location &31.00 & 29.00\\
 %\midrule
%$\text{Mistral-FT}_{\text{Atom}}$ & 12.00&20.00 \\
 %\quad \quad \quad w/o Location & & \\
%\midrule
$\text{Llama-3-FT}_{\text{Move+Copy}}$ &26.00 &27.00 \\
 \quad \quad \quad w/o Location &18.00 &22.00 \\
 \midrule
%$\text{Llama-3-FT}_{\text{Atom}}$ & 13.00&14.00 \\
 %\quad \quad \quad w/o Location & & \\
%\midrule
GPT-4o & 13.00& 15.00\\
 \quad \quad \quad w/o Location &11.00 & 14.00\\
\bottomrule
\end{tabular}
\caption{Acc (in percentage) of LLMs without location information. See~\tref{tab:location_plus} for Not M\% scores.}%\lemao{There are some exceptions in this table. Maybe you should readjust the prompts for w/o location.}

\vspace{-0.1in}
\label{tab:location}
\end{table}

\section{Details on crafting the small-size tasks in~\tref{tab:large matrix}}
\label{appendix:small size}
Specifically, we randomly initialized 100 tasks for Move and Copy from a range that is half of the original range listed in Appendix~\ref{appendix:araoc}. This results in 100 new tasks for both atomic operations, with an average size of $4.96 \times 4.89$ and $4.81 \times 4.80$, respectively. For comparison, the original average sizes are $10.07 \times 10.16$ and $9.72 \times 9.62$, respectively.

\section{Additional Results on banning the location information}
\label{appendix:banning}

To further study the effect of LLMs' understanding 
of matrix-format input, we evaluate two example LLMs that have strong performances on Move and Copy ($\text{Llama-3-FT}_{\text{Move+Copy}}$ in~\tref{tab:composition}, and GPT-4o) tasks in ARAOC since they could provide more reliable results. Specifically, we require them to finish the Move and Copy tasks again without using the location information of subgrids (the prompt is listed in~\fref{fig:without location prompt}), which should be important for finishing such tasks. 

As can be seen in~\tref{tab:location}, banning location information do significantly decrease these LLMs' performances on both tasks. These results again indicate that a fundamental understanding of matrices is crucial for completing ARAOC and ARC tasks.%The only exception appears on Copy with GPT-4o, which may be because GPT-4o is strong enough to infer transformation rules only using relative distances between two subgrids in the same grid. 

\section{Additional Results of~\tref{tab:autoregressive}}
\label{appendix:additional table12}

\begin{table}[tb]
\small
\centering
\setlength{\tabcolsep}{3.5mm}
\begin{tabular}{l|ll}
\toprule
\textbf{Direction}& \textbf{Mistral-8*7B}& \textbf{Llama-3-70B} \\
\midrule[0.5pt]
Left & 0.00 (73.00) & 11.00 (55.00)\\

Right & 7.00** (74.00)& 17.00 (49.00)\\

\bottomrule
\end{tabular}
\caption{Acc (in percentage) of LLMs with two mirroring directions. Not \% scores are listed in brackets.``**'' means 
 the bottom result is significant better than the upper one with p < 0.05.}
%\vspace{-0.2in}
\label{tab:autoregressive_addition}
\end{table}

In this section, we present the results of Mistral-8*7B and Llama-3-70B in~\tref{tab:autoregressive}. As shown, both LLMs continue to perform better when the mirroring direction is to the right, with a significant difference observed for Mistral-8*7B. This further reinforces the findings in~\tref{tab:autoregressive}.

\section{Additional Saliency Analysis Example}
\label{appendix:additional saliency}

\begin{figure*}
    \centering
    \includegraphics[width=0.96\textwidth]{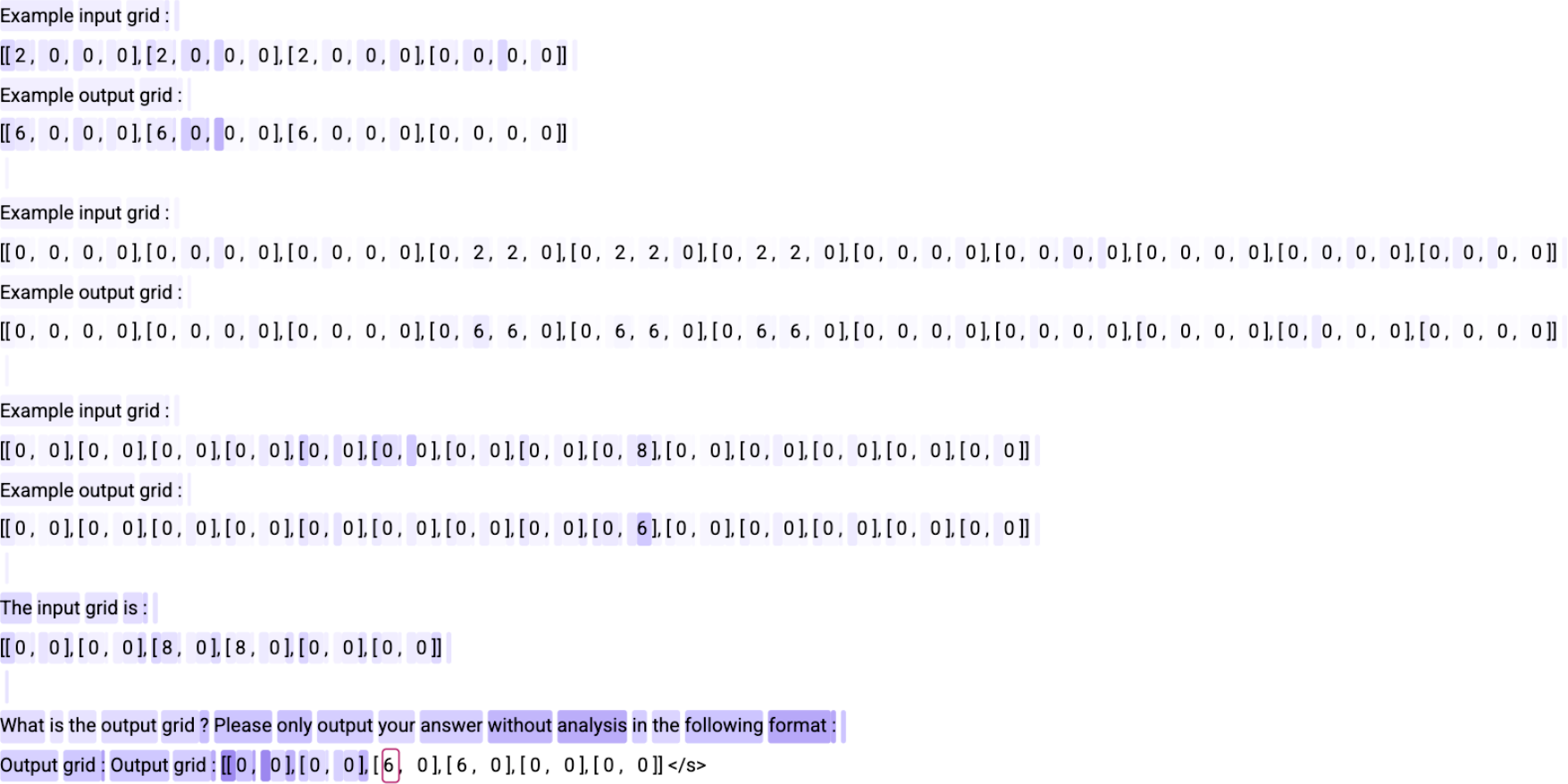}
    \caption{A saliency analysis example where Mistral makes a correct prediction. Darker means higher saliency corresponds to the boxed token. As can be seen, 
} 
\vspace{-0.2in}
    \label{fig:saliency_new}
\end{figure*}

In this section, we present an additional saliency analysis example in~\fref{fig:saliency_new}, where Mistral correctly predicts a Change Color task. As shown, Mistral not only accurately focuses on the "8" that needs to be changed in the testing input grid but also pays sufficient attention to the other modified parts in the in-context examples. This allows it to gather enough information about the task requirements, and finally leading to the correct prediction.

\section{Generalization of Our Findings}
\label{appendix:generalization}

In sections~\sref{sec: atom operation}, \sref{sec:factor}, \sref{sec:matrix}, \sref{sec:model}, we conclude several findings on LLMs’ fluid intelligence. In this section, we further discussing how can our findings generalize to other real-world tasks.

\begin{enumerate}
    \item \textbf{LLMs is largely affected by the superficial properties of the input, and fail to grasp the underlying concept of the operations.} LLMs often focus on superficial input properties and fail to understand the underlying concepts of operations. In real-world tasks like code generation, this manifests as a tendency to replicate syntax patterns from the input without looking for deeper logical relationships. For example, LLMs might generate syntactically correct but semantically incorrect code, similar to their reliance on superficial features in ARC and ARAOC tasks.

    \item \textbf{Fine-tuning does not teach LLMs how to induct operations from the in-context examples.} While fine-tuning can improve LLMs' performance on specific tasks (e.g., completing function templates or implementing algorithms in code generation), it does not enable LLMs to inductively generalize from in-context examples to unseen scenarios. For example, in code generation, after fine-tuning on demands requiring the KMP algorithm, an LLM might still struggle to apply the KMP algorithm to novel demands, reflecting its challenges with generalization in ARC and ARAOC.

    \textbf{LLMs' limitations on obtaining global representations (also partly due to the auto-regressive generation characteristic) of the input tasks affect their fluid intelligence.} LLMs' limitations in forming global representations, partly due to their autoregressive generation nature, also impact their performance in real-world tasks. In real-world LLM tasks especially when the input context is long, LLMs often fail to maintain consistent naming (e.g., variable and function names in code generation) or follow through on multi-step logical dependencies when processing the input. This is similar to their inability to compose atomic operations into a holistic solution in ARC and ARAOC tasks.
\end{enumerate}

\section{Additional Not M\% Results}

For simplicity, we do not list the Not M\% scores for several tables. Here we list the Not M\% scores for these tables in~\tref{tab:controllable_plus}, \tref{tab:large matrix_plus}, \tref{tab:composition_plus}, \tref{tab:fine-tune arc performance_plus}, \tref{tab:location_plus}, \tref{tab:natural language input_plus}, and~\tref{tab:autoregressive_plus}.

\begin{table}[tb]
\small
\centering
\setlength{\tabcolsep}{3mm}
\begin{tabular}{llcc}
\toprule
& \textbf{COMB}& \textbf{Llama-3} & \textbf{GPT-4o} \\
\midrule[0.5pt]
\multirow{6}{*}{\textbf{Move}}&Up 1 &30.00&2.00\\
&Up 2 &24.00&1.00\\
&Up 3 &28.00&2.00\\
\cmidrule{2-4}
&Up-right 1 &16.00&3.00\\
&Up-right 2 &26.00&1.00\\
&Up-right 3 &28.00&0.00\\
\midrule[0.5pt]
\multirow{6}{*}{\textbf{Copy}}&Up 1&6.00&2.00\\
&Up 2 &14.00&1.00\\
&Up 3 &16.00&0.00\\
\cmidrule{2-4}
&Up-right 1 &12.00&0.00\\
&Up-right 2 &14.00&1.00\\
&Up-right 3 &20.00&1.00\\
\bottomrule
\end{tabular}
\caption{Not M\% scores for~\tref{tab:controllable}.}

\label{tab:controllable_plus}
\end{table}

\begin{table}[tb]
\small
\centering
\setlength{\tabcolsep}{1.3mm}
\begin{tabular}{ll|cccc}
\toprule
& \textbf{Setting} & \textbf{Mistral} & \textbf{Llama-3} & \textbf{GPT-3.5} & \textbf{GPT-4o} \\
\midrule[0.5pt]
\multirow{2}{*}{\textbf{Move}}& Ori &36.00 &19.00 &27.00 &0.00 \\
& Small & 8.00&14.00 &1.00 &0.00\\
\midrule
\multirow{2}{*}{\textbf{Copy}}&Ori &43.00 &13.00 &29.00 &0.00\\
&Small & 13.00& 10.00& 1.00&0.00 \\
\bottomrule
\end{tabular}
\caption{Not M\% scores for~\tref{tab:large matrix}.}

\label{tab:large matrix_plus}
\end{table}

\begin{table}[tb]
\small
\centering
\setlength{\tabcolsep}{2mm}
\begin{tabular}{lcc|cc}
\toprule
\textbf{LLM} &\textbf{Move} & \textbf{Copy} & \textbf{Comp} %& \textbf{Rank}
\\
\midrule
$\text{Mistral-FT}_{\text{Move}}$ &2.00&34.00&44.00  \\
$\text{Mistral-FT}_{\text{Copy}}$ &6.00&12.00&7.00\\
$\text{Mistral-FT}_{\text{Move+Copy}}$ &6.00&6.00&15.00\\
\midrule
$\text{Llama-3-FT}_{\text{Move}}$ &9.00&8.00&11.00\\
$\text{Llama-3-FT}_{\text{Copy}}$ &5.00&1.00&11.00\\
$\text{Llama-3-FT}_{\text{Move+Copy}}$ &6.00&1.00&3.00\\
\midrule
GPT-3.5 &27.00&29.00&42.00\\
GPT-4o &3.00&6.00&1.00\\
\bottomrule
\end{tabular}
\caption{Not M\% scores for~\tref{tab:composition}}

\label{tab:composition_plus}
\end{table}

\begin{table*}[tb]
\renewcommand\arraystretch{1.1}
\centering
\setlength{\tabcolsep}{2.5mm}
\small
\begin{tabular}{lcccccc|c}
\toprule[1pt]
\multirow{2}*{LLM} & \multicolumn{6}{c}{\textbf{Individual Atomic Operation}} & \multicolumn{1}{c}{\textbf{ Composition}} \\
\cmidrule{2-8}
& \multicolumn{1}{c}{\textbf{Move}} & \multicolumn{1}{c}{\textbf{Change Color}} & \multicolumn{1}{c}{\textbf{Copy}} & \multicolumn{1}{c}{\textbf{Mirror}} & \multicolumn{1}{c}{\textbf{Fill Internal}} & \multicolumn{1}{c}{\textbf{Scale}} & \multicolumn{1}{c}{\textbf{ARC}}  \\
% & Acc$\uparrow$ & Acc$\uparrow$ & Acc$\uparrow$ & Acc$\uparrow$ & Acc$\uparrow$ & Acc$\uparrow$ & Acc$\uparrow$ \\
\midrule[0.5pt]
%Mistral &36.00&30.00&43.00&97.00&31.00&98.00&48.00 \\
%$\text{Mistral-FT}$  &11.00&0.00&6.00&52.00&2.00&0.00&38.00 \\
%%\midrule
Llama-3 &19.00&17.00&13.00&96.00&6.00&89.00&33.00  \\
$\text{Llama-3-FT-arc}$ &20.00&14.00&7.00&92.00&1.00&95.00& 29.00\\
$\text{Llama-3-FT-atomic}$ &9.00&1.00&8.00&54.00&1.00&4.00&40.00  \\
$\text{Llama-3-FT-atomic-arc}$ &14.00&0.00&10.00&39.00&0.00&6.00& 30.00 \\
\midrule
%GPT-3.5 &27.00&13.00&29.00&89.00&12.00&80.00&35.00 \\
%GPT-4 &3.00&0.00&6.00&52.00&0.00&70.00&16.00  \\
GPT-4o&0.00&0.00&0.00&48.00&0.00&72.00&11.00 \\
\bottomrule[1pt]
\end{tabular}
\caption{Not M\% scores for \tref{tab:fine-tune arc performance}.}
\vspace{-0.2in}
\label{tab:fine-tune arc performance_plus}
\end{table*}

\begin{table}[tb]
\small
\centering
\setlength{\tabcolsep}{4mm}
\begin{tabular}{l|cccc}
\toprule
LLM& Move & Copy \\
\midrule
%$\text{Mistral-FT}_{\text{Move+Copy}}$ &25.00 &32.00 \\
 %\quad \quad \quad w/o Location &31.00 & 29.00\\
 %\midrule
%$\text{Mistral-FT}_{\text{Atom}}$ & 12.00&20.00 \\
 %\quad \quad \quad w/o Location & & \\
%\midrule
$\text{Llama-3-FT}_{\text{Move+Copy}}$ &6.00&1.00 \\
 \quad \quad \quad w/o Location &6.00&0.00\\
 \midrule
%$\text{Llama-3-FT}_{\text{Atom}}$ & 13.00&14.00 \\
 %\quad \quad \quad w/o Location & & \\
%\midrule
GPT-4o &0.00&0.00\\
 \quad \quad \quad w/o Location &0.00&1.00 \\
\bottomrule
\end{tabular}
\caption{Not M\% scores for~\tref{tab:location}.}

\vspace{-0.1in}
\label{tab:location_plus}
\end{table}

\begin{table}[tb]
\small
\centering
\setlength{\tabcolsep}{0.5mm}
\begin{tabular}{ll|cccc}
\toprule
& \textbf{Method}& \textbf{Mistral}& \textbf{Llama-3} & \textbf{GPT-3.5}& \textbf{GPT-4o} \\
\midrule[0.5pt]
\multirow{2}{*}{\textbf{Move}}& w/o NL &-&-&-&-\\
& NL &82.00&0.00&0.00& 1.00\\
\midrule[0.5pt]
\multirow{2}{*}{\textbf{Color}}& w/o NL &-&-&-&- \\
& NL &83.00&0.00&1.00&0.00\\
\midrule[0.5pt]
\multirow{2}{*}{\textbf{Copy}}& w/o NL &-&-&-&- \\
& NL &83.00&4.00&2.00&0.00 \\
\midrule[0.5pt]
\multirow{2}{*}{\textbf{Mirror}}& w/o NL &-&-&-&- \\
& NL &80.00&30.00&27.00&0.00\\
\midrule[0.5pt]
\multirow{2}{*}{\textbf{Fill Internal}}& w/o NL &-&-&-&-\\
& NL &84.00&1.00&0.00&0.00\\
\midrule[0.5pt]
\multirow{2}{*}{\textbf{Scale}}& w/o NL &-&-&-&-\\
& NL &98.00&75.00&63.00&41.00 \\
\bottomrule
\end{tabular}
\caption{Not M\% scores for~\tref{tab:natural language input}. For scores under ``w/o NL'', please refer to~\tref{tab:araoc results}.}
\vspace{-0.1in}
\label{tab:natural language input_plus}
\end{table}

\begin{table}[tb]
\small
\centering
\setlength{\tabcolsep}{2mm}
\begin{tabular}{l|cccc}
\toprule
\textbf{Direction}& \textbf{Mistral}& \textbf{Llama-3} & \textbf{GPT-3.5}& \textbf{GPT-4o} \\
\midrule[0.5pt]
Left &68.00&76.00&52.00&33.00\\

Right &71.00&74.00&49.00&22.00\\

\bottomrule
\end{tabular}
\caption{Not M\% scores for~\tref{tab:autoregressive}.}
\vspace{-0.1in}
\label{tab:autoregressive_plus}
\end{table}